**White Paper**

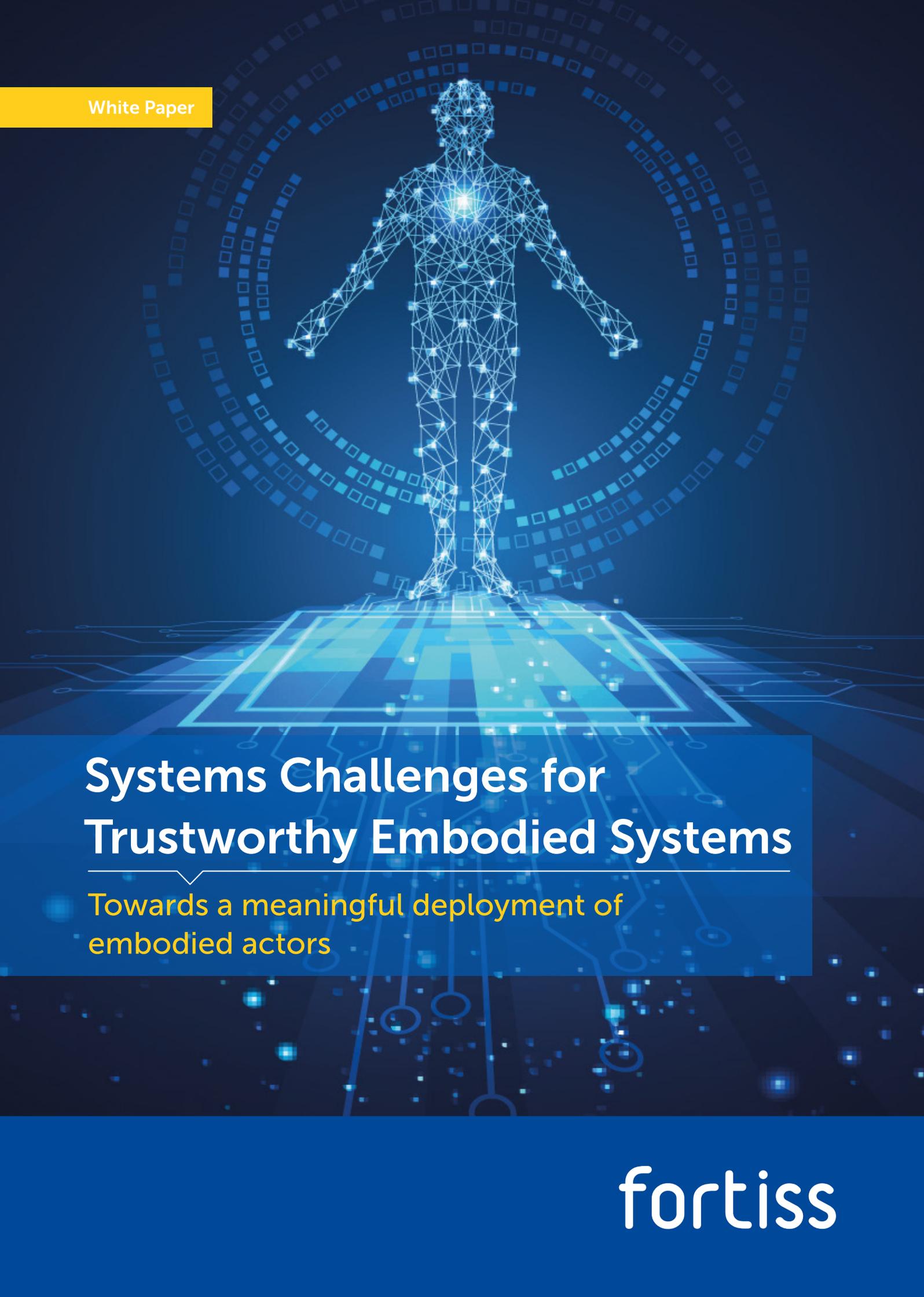

# Systems Challenges for Trustworthy Embodied Systems

## Towards a meaningful deployment of embodied actors

**fortiss**

# Systems Challenges for Trustworthy Embodied Systems –
Towards a meaningful deployment of embodied actors

Autor

**Dr. Harald Rueß**

*fortiss GmbH*
*Guerickestr. 25*
*80805 München*

ruess@fortiss.org

# Systems Challenges for Trustworthy Embodied Systems[1]
## Towards a meaningful deployment of embodied actors

Autor:

Dr. Harald Rueß

fortiss - Research Institute of the Free State of Bavaria

Guerickestr. 25

80805 Munich

Version 2.0

15th of March, 2022

This research is supported by the BMWi-funded project Embodied Intelligence - The Next Big Thing, and the CASSAI project as funded by the Bavarian Ministry of Economics in the context of the fortiss AI Center.

Supported by:

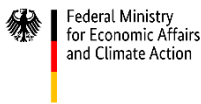

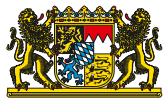

Bavarian Ministry of Economic Affairs, Regional Development and Energy

---

[1] Disclaimer: This research is supported by project Embodied Intelligence—The Next Big Thing as funded by the German Ministry of Economics, and the CASSAI project as funded by the Bavarian Ministry of Economics in the context of the fortiss AI Center. This report heavily draws on the results of the Agenda CPS and ongoing discussions, and constructive feedback on earlier versions by Prof. Dr. Dr. h.c. Manfred Broy, is strongly influenced by ongoing work by SRI colleagues on the rigorous design of increasingly autonomous machines, in particular Dr. John Rushby and Dr. Natarajan Shankar, and also exchanges during a research stint in November 2019. It is also based on regular exchanges with fortiss colleagues, in particular, Dian Balta, Dr. Markus Duchon, Dr. Johannes Kroß, Prof. Dr. Daniel Mendez, Prof. Dr. Rute Sofia, and Dr. Henrik Putzer. It has also benefited from a recent BIG workshop on challenges for embodied systems with participating colleagues from Siemens and fortiss. The views in the section on robust AI were developed in 2017 in preparation of the fortiss-IBM joint research center, and the section on human-centered engineering is heavily based on the corresponding fortiss whitepaper by Dr. Yuanting Liu and Dr. Habil. Hao Shen on this topic. The views on uncertainty quantification were mainly formed through interactions within the CASSAI projects with the fortissians Carmen Carlan, Amit Sahu, Tewodros Beyene, and Julian Bernhard. It is a safe assumption, however, that most probably none of the contributors mentioned above could agree with all the described hypotheses on systems challenges for embodied systems as outlined here.

# Abstract

A new generation of increasingly autonomous and self-learning systems, which we call embodied systems, is about to emerge. When deploying these systems into our very societal fabric, we face various engineering challenges, as it is crucial to coordinate the behavior of embodied systems in a beneficial manner, ensure their compatibility with our human-centered social values, and design verifiably safe and reliable human−machine interaction. We argue that traditional systems engineering is coming to a climacteric from embedded to embodied systems, and with assuring the trustworthiness of dynamic federations of situationally aware, intent-driven, explorative, ever-evolving, largely unpredictable, and increasingly autonomous embodied systems in uncertain, complex, and unpredictable real-world contexts.  With this goal in mind we identify urgent systems engineering challenges for designing embodied systems in which we can put our trust, including robust and human-centered artificial intelligence, cognitive architectures, uncertainty quantification, trustworthy self-integration, and continual analysis and assurance.

# Table of Contents

# Introduction

A new generation of increasingly autonomous machinery is about to be developed and embodied into all aspects of everyday life. This machinery is used beyond mere automation and assistance to humans. As manufacturing robots make way for autonomous machine workers, business and administrational services are performed by autonomous virtual organizations, and processes and value chains in the real and virtualized worlds are executed by coalitions of autonomous machine actors. In this way, computational machines act as self-sufficient entities in our very economic and societal fabric.

Next-generation autonomous machines may also acquire and improve necessary skills and behavior in real-world contexts by means of action, reaction, and interaction in and with physical and social environments.[2] But how exactly do the required cognitive skills and behavior that could possibly be termed *intelligent*[3] unfold in time?

Various answers to this fundamental quest on the emergence of intelligence, including Descartes' brain–body duality, have been developed throughout the history of philosophy. These considerations have also been motivating the field of *embodied intelligence (EI)*, which is the *computational approach to the design and understanding of intelligent behavior in embodied and situated actors through the consideration of the strict coupling between the actor and its environment (situatedness), mediated by the constraints of the actor's own body, perceptual and motor system, and brain (embodiment).*[4]

On the basis of this paradigm, computational machines may develop and improve at least some physical dexterities and cognitive skills by means of interactions between the physical *self* and environment.[5] [6] [7] [8] [9] [10] [11] [12] [13] [14] In this way, EI robots that can move, see, speak, and interact with

---

[2] Striedinger. *Hauser, Kaspar, der "rätselhafte Findling" 1828–1833, Lebensläufe aus Franken, Herausgegeben im Auftrag der Gesellschaft für Fränkische Geschichte von Anton Chroust, III*. Erster Band, 1927.
[3] Deary. *Intelligence: A Very Short Introduction*. Oxford University Press, 2020.
4 Cangelosi, Bongard, Fischer, Nolfi. *Embodied Intelligence*. Handbook of Computational Intelligence. Springer, 2015.
[5] Growing evidence shows that at least some sensorimotor behavior and cognitive skills are realized in the body and that social interactions can bootstrap learning (for example, Ballard et al., *Deictic codes for the embodiment of cognition*. Behavioral and Brain Sciences, 1997; McGeer. *Passive dynamic walking*. International Journal of Robotics Research, 1990; O'Regan, Noe. *A sensorimotor account of vision and visual consciousness*. Behavioral and Brain Sciences, 2001)
[6] Pfeifer, Scheier. *Understanding Intelligence*, MIT Press, 2001.
[7] Breazeal. *Designing Sociable Robots*. MIT Press, 2002.
[8] Beer. *A dynamical systems perspective on agent-environment interaction*. Artificial Intelligence 72, 1994.
[9] Brooks. *Elephants don't play chess*, Robotics and Autonomous Systems 6(1), 1990.
[10] Cangelosi. *Grounding language in action and perception: From cognitive agents to humanoid robots*. Physics of Life Reviews 7(2), 2010.
[11] Chiel, Beer. *The brain has a body: Adaptive behavior emerges from interactions of nervous system, body and environment*. Trends in Neurosciences 20, 1997.
[12] Keijzer. *Representation and Behavior*. MIT Press, 2001.
[13] Nolfi, Floreano. *Evolutionary Robotics: The Biology, Intelligence, and Technology of Self-Organizing Machines*. MIT Press, 2000.
[14] Pfeifer, Bongard. *How the Body Shapes the Way We Think: A New View of Intelligence*. MIT Press, 2006.

other robots effectively have been built.[15] [16] Current approaches to EI also rely on training actors in virtualized playgrounds for accelerated learning and for enabling nondestructive learning from failure.[17] [18] [19] [20] Despite all this progress on EI, however, it still remains open to speculation if and how "higher-level" thinking, symbol grounding, natural language understanding, consciousness, and emotions may emerge in machines through embodiment.[21] [22] [23]

A central but largely untouched challenge of EI is concerned with embodying computational actors into real-world environments with all their intricacies, including uncertainty, complexity, and unpredictability. This is not an easy task as actors may demonstrate unexpected and even emergent behavior when deployed in new operating environments, such as when being moved from a model-based virtual playground into a real-world physical context.[24]

Another key point of moving actors from innocent training playgrounds into a physical or social context is that actions have real consequences. Tay,[25] a pretrained chatbot that quickly turned nasty, when placed in the context of a (so-called) *social network*, may serve as a cautionary tale of what can go wrong all too easily and quickly. While this problem was solved by simply taking Tay offline, it is easy to imagine other cases of self-learning actors neglecting human moral expectations with a long-lasting and perhaps even more severe impact. The sad tale of Tay also reminds us of the importance of meaningful control of EI.

We clearly face serious social, economic, and legal challenges when deploying EI in the real world,[26] as it is crucial to coordinate the behavior of EI actors in a *beneficial* manner, ensure their compatibility with our human-centered *social norms* and *values*, and design verifiably safe and reliable human–machine interaction.

Notice that interpretations of the terms "beneficial" and "social norms and values" are heavily context- and application-specific. A simple robot companion, for example, might be called beneficial if it acts according to the intents of its human companion. This idea does not necessarily imply, however, that the robot companion acts according to the intents of others, which may even be contradictory, let alone that it acts according to relevant human-centered social values.

---

[15] Workshop on Embodied AI at CVPR'21 (https://embodied-ai.org)
[16] Pfeifer, Iida. *Embodied Artificial Intelligence*. Springer, 2004.
[17] Such as visual exploration, visual navigation, and embodied question-answering. In particular, the question "Is there still milk in the fridge?" may require the embodied actor to unlock new insights that are momentous in answering and initiate corresponding tasks such as moving to the kitchen and opening the fridge.
[18] For example, workshop on Embodied AI. http://https://embodied-ai.org/. CVPR, 2020.
[19] For example, ongoing Embodied AI programs, such as at Intel Labs or Facebook AI Research.
[20] These simulation environments currently include SUNCG, Matterport3D, iGibson, Replica, Habitat, and DART.
[21] Smith, Gasser. *The development of embodied cognition: Six lessons from babies*. Artificial Life, 11(1–2), 2005.
[22] Rushby, Sanchez. *Technology and Consciousness*. SRI Technical Report, 2018.
[23] For example, integrated information theory (IIT) asserts that it is the intrinsic causal powers of the brain that really matter. Those powers cannot be simulated but must be part and parcel of the physics of the underlying mechanism.
[24] Remark: The very notion of a "digital twin" suggests an unequivocalness of the model with respect to the real world no model can actually deliver; as Plato already realized, such a model is nothing more than a "shadow on the wall."
[25] https://www.theverge.com/2016/3/24/11297050/tay-microsoft-chatbot-racist
[26] Vazdanan et al. *Responsibility Research for Trustworthy Autonomous Systems*. 2021.

In our quest to develop and deploy EI in a meaningful way, we need to find adequate solutions to central engineering challenges:

- How can we ensure that an EI actor behaves beneficially? That is, it functions as intended and behaves in accordance with higher-level societal goals and standards.
- How can we ensure that learning-enabled EI actors are robust across their whole life cycle? That is, the behavior of EI-enabled actors is dependable, safe, and predictable (up to quantified tolerances) in uncertain, complex, and unpredictable environments.
- How can we ensure that meaningful (human) control over an EI actor is enabled during operation?

Traditionally, the field of *systems engineering* tackles these kinds of questions for assuring purposeful and acceptable technical systems.

However, the engineering of software-intensive systems has so far mainly been concerned with relatively small-scale, centralized, deterministic, nonevolvable, automated, and task-specific *embedded* and *cyber-physical systems* (CPSs), which operate in well-defined and largely predictable operating environments. The state-of-the-practice in *safety engineering*,[27] for instance, is restricted to deterministic systems operating in well-defined operating contexts, and it usually relies on ultimate fallback mechanisms to a human operator (as is the case, for example, for current airplane autopilots). Current safety engineering practice, therefore, does not support certification and corresponding operating readiness of the envisioned new generation of EI acting in real-world environments.

In Section 2, we introduce the notion of *embodied systems.* This class of systems is based on the central EI concepts of situatedness and embodiment. As such, the design of embodied systems may benefit from any progress on EI, but their success does not hinge on reaching EI's ultimate quest for human-like machine intelligence.

---

[27] For instance, industrial safety engineering standards, such as DO 178C in aerospace and ISO 26262 in the automotive industry.

We illustrate the defining features and disruptive potential of embodied systems by means of three different scenarios, namely, software assistants, robot companions, and federations of services. In Section 3, we analyze the main characteristics of embodied systems, and in Section 4, we discuss how to possibly increase trust in embodied systems. In Section 5, we deduce system challenges for designing trustworthy embodied systems from the main characteristics of embodied systems. We conclude with some final remarks in Section 6.

# 1. Embodied Systems

In contrast to traditional EI, which is largely driven by its fundamental quest for emerging intelligent behavior in computational machines, we are embracing a seemingly utilitarian stance in that the embodiment of actors in a real-world context, physical or social, serves

- the development of flexible, inventive, and optimized actions
- for beneficial, goal-oriented, and robust behavior
- in uncertain, complex, and unpredictable real-world environments.

The central challenge is to enable the beneficial, dependable, safe, and predictable—at least up to acceptable tolerances—operation of embodied actors when woven into our very economic and social fabric.

Actors are supposed to purposefully cooperate with other actors, both human and robot. As such, embodied systems support humans in various real-world tasks and in an increasingly autonomous, self-guided manner. Embodied systems, however, should not be restricted to an assistance role. Instead, their self-learning capabilities should well support the coevolution of humans and machine as necessary for bootstrapping overall capabilities.[28]

*Embodied systems*[29] may be viewed as increasingly autonomous and self-learning CPSs.[30] More specifically, embodied systems are decentralized, dynamic, self-learning, and self-organizing federations of actors that collaborate to accomplish complex tasks and missions in partially unpredictable real-world operating contexts. Therefore, these systems are:

- **Situational** in that they are aware of the operating context and itself.
- **Embodied** in that real-world actions, reactions, and interactions are instrumental for the flexible development of inventive actions for goal-oriented and robust behavior in uncertain, complex, and unpredictable real-world environments.
- **Open** to interacting and collaborating with others in a mutually synergistic manner while still operating as self-sufficient individually purposeful systems.

---

[28] Bardini. *Bootstrapping: Douglas Engelbart, Coevolution, and the Origins of Personal Computing*. Stanford University Press, 2000.
[29] Notice that we use the terms *embodied system* and *embodied actors* almost interchangeably; if the embodied system is regarded as a self-sufficient component, however, we tend to call it an actor. Moreover, populations of actors are usually referred to as embodied systems.
[30] Broy, Geisberger (Eds.). Agenda CPS. acatech, 2013 (https://www.fortiss.org/en/results/scientific-publications/details/agenda-cps-integrierte-forschungsagenda-cyber-physical-systems).

- **Adaptive** in that they may adjust and improve behavior through experience and targeted exploration.

Some scenarios might help illustrate the main concepts, benefits, and challenges of embodied systems.

## Software Actors

Consider a mission- and communication-oriented software-based actor[31] that autonomously gathers information, for example, from seminar organizers, composes announcements of next week's seminars, and mails them each week to a list that it keeps updated, all without human supervision. Similarly, consider a software companion for automated assignment of tasks to employees in a large organization that automates the workload of hundreds of human counterparts. Automated software assistants typically are "embodied" in and interact with a largely virtual world of information sources, with a dedicated interface to a human operator. These kinds of autonomous software assistants do exist, with their main architectural principles rooted in cognitive architectures such as global workspace theory[32] or multiactor communication frameworks (KQML[33] or OAA[34]). They have often been shown to outperform human counterparts on many real-life exploration and complex planning, scheduling, and dispatching tasks such as customer service, risk evaluation, product inspection, and data mining. Their actions are based on a situational awareness of the operating environment together with built-in intents and goals for managing the unpredictable.

Traditional autonomous software assistants, however, lack the main attributes of embodied systems as they are purpose-built for solving specific tasks only, and they have a limited ability to adapt to changing environments through interactions with their environment, including human operators.

We can easily imagine a new generation of integrated and more versatile software-based actors, which are embodied in the sense that they are learning our intents through collaborative interactions and are able to automatically detect and adapt to changing conditions (for example, a specific source of information is temporarily unavailable), thereby also optimizing future problem-solving strategies. Such a personal software assistant should be able to largely automate managerial tasks, such as doing taxes, planning trips, ordering food, or controlling household appliances, according to an overall situational assessment and intents.

---

[31] The actor paradigm in AI is based upon the notion of reactive (internally motivated entities embedded in changing) uncertain worlds that they perceive and in which they act.
[32] Franklin, Patterson. *The LIDA Architecture: Adding New Modes of Learning to an Intelligent, Autonomous, Software Agent. IDPT, 2006.*
[33] Finin, Labrou, Mayfield. *KQML as an Agent Communication Language.* Software Agents, MIT Press, 1997.
[34] Martin, Cheyer, Moran. *The open agent architecture: A framework for building distributed software systems.* Applied Artificial Intelligence, 13(1–2), 1999.

## Robotic Companions[35]

Figure 1. depicts an embodied system with the sole purpose of landing an airplane. In a slightly more complex scenario, we envision such a robot to act as a copilot in a single-pilot cockpit. In this way, on long flights with two pilots, one can sleep while the other flies with assistance from the robotic companion. Such a robot needs to be more like a human copilot than a conventional flight management system or a functionally automated autopilot. In particular, the robot companion needs to perform heterogeneous and complementary tasks, including radio communications, interpreting weather data and making route adjustments, pilot monitoring tasks, shared tasks (flaps, gear), ground taxi, and communication with the cabin crew (emergency evacuation).

The robot companion also needs to integrate these tasks to accomplish a safe flight; it needs to base its decisions and actions on an overall situational assessment. In case things go wrong, the robot companion needs to find effective explanations based on fault diagnosis and engage in an effective resolution process with the (human) pilot on the basis of a model of the pilot's beliefs.

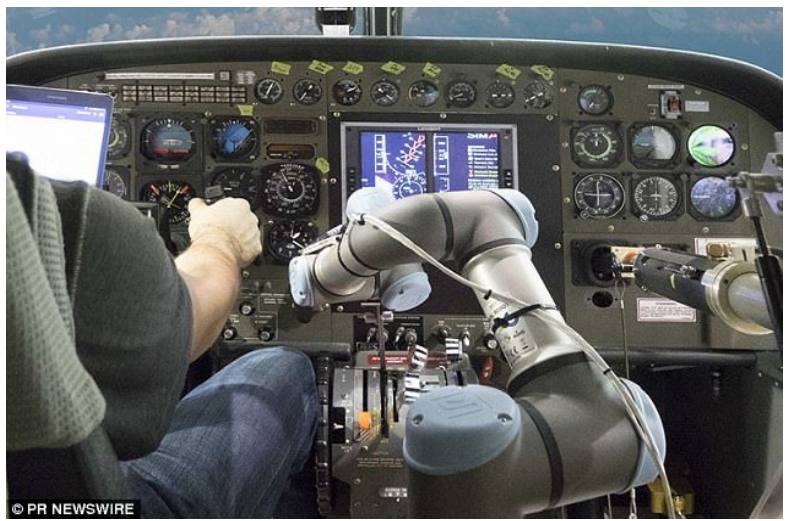

*Figure 1. Robotic copilot.[36]*

---

[35] This scenario is heavily based on the one presented by J. Rushby on increasingly autonomous systems.
[36] Source: https://www.aerotime.aero/23374-robot-co-pilot-flies-lands-boeing-737

In extreme situations, the automated robot copilot might even take over control, such as if there is smoke in the cockpit. The robotic copilot might also be better suited to maintain pitch and thrust in extreme situations as the human pilot.[37] In these rare cases, the "robot" must now also cope with inconsistencies (for example, in sensor readings) based on flight laws, training procedures, models of the physical environment, and unforeseen situations without the possibility of a structured handover to the human pilot. In extreme cases where flight laws suddenly change, due, for example, to severe damage to the body of the airplane, the mechanized copilot can even relearn to fly the aircraft under these new circumstances on the spot and in real time.

Altogether, the robotic copilot in Figure 1. augments the sensing, acting, and mental capabilities of the pilot (Figure 2). For its embodied "self," it may be operated in different cockpit settings. However, if the robotic copilot is only supposed to augment the decision-making and acting capabilities of a pilot, then an embodied "self" is superfluous. In these cases, deploying an autonomous and self-learning software companion into the cockpit control system is sufficient.

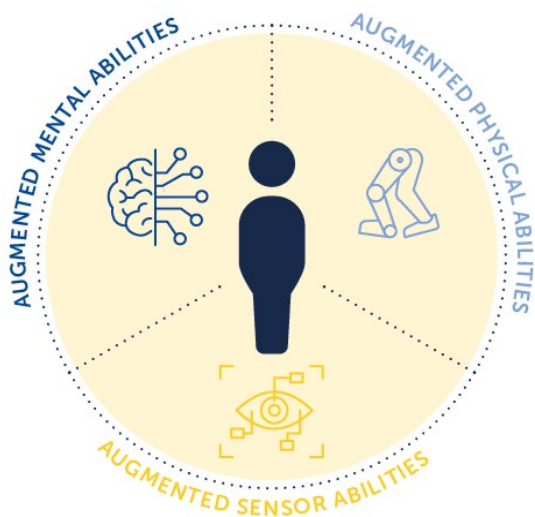

*Figure 2. Areas of human augmentation.*

The robotic copilot is a particular instance of a larger class of embodied companions. Personal companions for supporting and taking over tedious household chores and for assisting with tax declarations, including communication with tax authorities, and suggesting new possibilities based on our intents are an old dream.

---

[37] For instance, http://understandingaf447.com/extras/18-4_minutes__23_seconds_EN.pdf

Embodied companions are also designated to assist truck drivers, ship captains, caregivers, investors, managers, workers, lawyers, and, in fact, everybody. Potential benefits of these kinds of embodied companions include increased safety, reliability, efficiency, affordability, and previously unattainable capabilities.

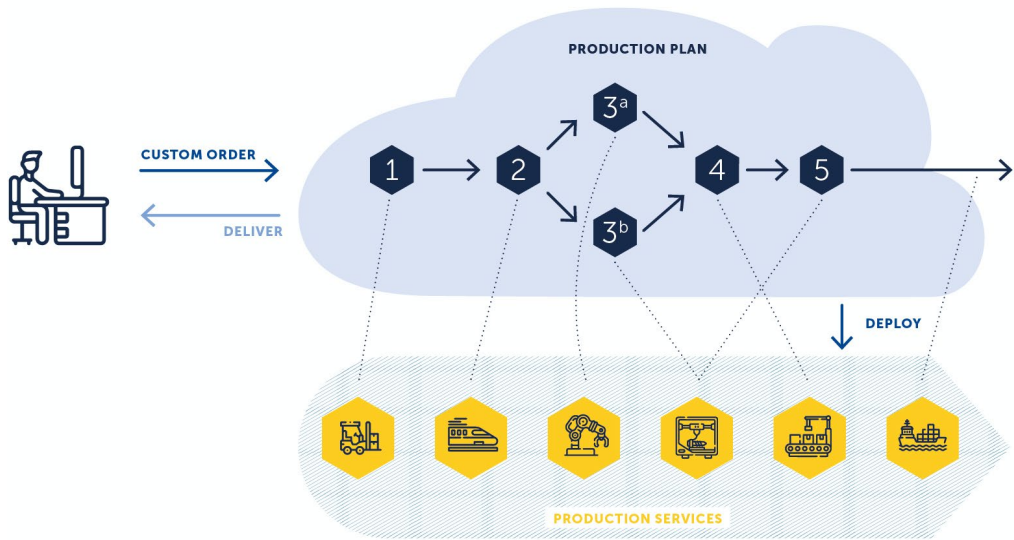

*Figure 3. Federated retail.*

## Service Federations

When using current shopping websites, the consumer selects an order from a large but ultimately fixed set of product options and configurations, and the ordered goods are, by and large, already produced and stored in a system of warehouses to guarantee speedy delivery. As such, the traditional shopping experience has just moved from a physical store to a retail website, but the whole process builds on established stakeholder roles of consumers, producers, logisticians, and chains of retailers.

We might think, however, of more flexible setups that start with a software-defined custom design instead of a custom order. An often-cited example is custom-designed sneakers that might even be 3D knitted, say, in a store.[38] Other obvious examples are software-defined custom cars or mission-specific drones. The journey thus starts with a custom design, which is usually constructed by instantiating available design patterns and composing them into the desired design and a corresponding production plan.

---

[38] See Speed Factory (https://www.digitale-technologien.de/DT/Redaktion/DE/Standardartikel/AutonomikFuerIndustrieProjekte/autonomik_fuer_industrie_projekt-speedfactory.html)

This new design and the production plan can be made available to others to build upon, analogous to an app store. Let us assume, however, that the design now needs to be realized, that is, the production plan needs to be executed. Instead of identifying a suitable production site upfront, a bidding process is started among several production sites and logistics providers to dynamically set up a suitable distributed production chain for realizing the custom design.

In Figure 3, this process of connecting the digital production plan with real-world production entities, such as 3D printers, robots, and flexible production lines together with storage and transport capabilities, for moving partly finished products is referred to as *deployment*.[39] Some of these entities, such as the shipping service, may operate autonomously and are possibly organized as their own virtual companies.

Such a deployment might be chosen according to different criteria, including the traditional triad of costs, quality, and time-to-delivery. For example, the ordering customer is provided with the option of a reduced price if they accept a, say, two-day shipping delay. Additionally, flexible deployment might take other all-important attributes such as resilience to faults or even partial collapse of selected production or logistics capabilities, sustainability (for example, climate neutrality), and other ethical considerations (for example, no child labor). These attributes might now even be part of the ordering process itself, thereby enabling everybody to act on behalf of their own personal beliefs and social values and directly influence the world through seemingly mundane everyday acts such as ordering goods.

Altogether, the customer is not (only) a customer anymore because they can order the manufacturing of the goods they want or need. Moreover, integrated production and logistics chains are broken up into highly specialized services, which are orchestrated flexibly and in a highly distributed fashion. Production and logistics services might be realized through in-creasingly autonomously acting actors, who might do their own bidding for providing services with the purpose of forming dynamic coalitions. From this point of view, there does not seem to be a direct need for a traditional retailer[40] as an intermediary between customers and producers anymore.

By and large, we might be able to realize a proof-of-concept prototype of the embodied retail system as depicted in Figure with the currently available technology. In fact, the attentive reader might already have noticed that the embodied retailer is structured according to the main principles of *model-driven design*, namely, platform-independent model, platform-dependent modeland a deployment relation between those two models.

---

[39] In general, such a deployment is an n:m relation between services in the virtual world and their physical embodiments.
[40] Maybe as an embodied system?

Systems Challenges for Trustworthy Embodied Systems                                                                                                    13

However, the extra flexibility of the embodied retail system comes with a cost, and we do not yet have a foundational understanding of the necessary operators for composing the required designs and service nor do we have anything that comes close to the required large-scale deployment schemes. Serious questions have also been raised regarding resilience—that is, tolerance, isolation, and recovery from faults, breakdowns, and cyberattacks.[41]

From the business and societal perspective, the main open questions involve the feasibility of corresponding value chains, incentivization and regulation, (product) liability issues, intellectual property, payment systems, and tax laws. Indeed, systems, such as the outlined federation for retail, do not only have the potential to disrupt established economic cycles but also shake up established social, economic, and governmental structures and infrastructures.

The embodied retail scenario has been designed to also demonstrate the possible elimination of middlemen between consumer and service providers, thereby enabling completely new value chains in the upcoming era of *postplatform ecosystems*. These kinds of dynamic federations of embodied service systems may well make currently well-established retail platforms and middlemen, such as Amazon, Uber, or AirBnB, superfluous. They also rely on flexible infrastructural services for payment systems, insurance, and arbitration boards, among others.

We can easily abstract from the main building principles of embodied retail to other societal-scale lifeworlds to create similar scenarios; in particular, critical supply infrastructures (for example, energy/water), mobility (for example, integrated transportation), medical (for example, personalized medicine), law (for example, judicial Q&A advisors, attorneys-at-law, and software judges in line with Leibniz's rational actors), and government (for example, seamless and proactive government services).

Numerous other examples of embodied federated systems can be found across many application domains, such as *collaborative IoT* systems for supply–demand interaction (for example, machine–machine, human–machine) in industrial applications, prosumer interaction and collaboration under permanently changing conditions in the energy domain*, ad hoc* collaboration between smart infrastructure and autonomous vehicles, self-engineering and reconfiguration of automation components in building automation, or self-managed production logistics and warehouse logistics with a heterogeneous fleet of AGVs, which are able to self-adapt to heterogeneous production floors and tasks.

---

[41] Fault detection, isolation, recovery (FDIR)

Moreover, next-generation communication networks are expected to sense, compute, learn, reason, act on business intent, and manage the ongoing explosion of data from an ever-increasing number of connected intelligent devices and a multiplicity of new use cases, along with ever-changing network topologies.[42] Zer*o-touch* network operations for coping with the growing system's complexity, size, and reduced decision-making lead time result in an increased level of network autonomy and self-adapting automation, where unforeseen situations, intents, and requirements can usually be addressed without human intervention.

---

[42] Ericsson. *Artificial intelligence in next-generation connected systems*. Whitepaper, 2021.

## 2. Characteristics

Embodied systems comprise federations of collaborating actors. They operate in largely unpredictable environments, physical or not, and they recognize this environment through sensors. Moreover, they are informed about the intentions of other actors in their respective and immediate operating environments; take nontrivial decisions based on reasoning; influence their environment, including other actors, via actuators; interact and cooperate with the elements of their operating environment; influence elements in their environment to better meet their own goals[43]; show a certain behavior based on skills; and learn new and improved behavior during operation and through interactions.[44] Altogether, an embodied system is characterized as being:

- **Cognitive.** as actions are based on situational awareness, model building, and planning.
- **Intent-driven.** as actions are based on capturing actors' intents, tasks, and goals.
- **Federated.** as actions of decentralized actors are coordinated in a collaborative manner between stakeholders and on an intentional level to accomplish joint tasks or missions.
- **Autonomous.** as actions are increasingly determined by an actor's, or federations of actors', own knowledge, beliefs, intents, preferences, and choices.
- **Self-learning.** as actions are adapted and improved through experience, exploration, and reasoning, both inductive and deductive, of a situated actor.

Embodied systems may, therefore, be viewed as autonomous and self-learning CPSs with intent-driven, goal-oriented, and collaborative behavior.

Embodied systems, therefore, go well beyond current *functional automated* systems, such as autopilots for landing an airplane, parking a car, or manufacturing robots, which are all designed to handle limited tasks in precisely specified operational contexts. Task execution in functional automated systems is planned offline or during design time and they usually do not learn during operation. Moreover, collaboration is restricted to the exchange of information about the system context, and they rely on a fallback mechanism to a human operator when encountering unforeseen and difficult situations. Most current systems, including production or household robots, are functional automated systems, and we are on the edge of deploying intent- and mission-oriented systems. Simple collaborative systems have started to be established. Self-learning during operation, however, is currently impossible for mission-critical and safety-related applications.

---

[43] For example, Mechanism Design (see also: Roughgarden. Algorithmic Game Theory. CACM, 2010).
[44] Putzer, Wozniak. *Trustworthy Autonomous/Cognitive Systems*. fortiss, 2021. (see also: https://www.fortiss.org/fileadmin/user_upload/05_Veroeffentlichungen/Informationsmaterialien/fortiss_whitepaper_trustworthy_ACS_web.pdf)

## Cognitive Systems

Cognition may be defined as the mental action or process of acquiring knowledge and understanding through experience and the senses.[45] Clearly, it is only in conjunction with cognitive faculties that systems and machines develop their full potential and address a constantly growing number of new challenges in daily use.

In a system with cognitive faculties, individual or interconnected objects, therefore, cannot only perceive the physical environment but are also able to learn from the wealth of experience gained, derive new insights, understand contexts, and make important decisions in a supportive or autonomous manner.

The essential cognitive faculties are:[46] [47]

- **Perception.** Fusion and interpretation of a multitude of sensors, stimuli, and observed behavior; removal of vagueness and ambiguity in input data; and synthesis of relevant information such as the detection, localization, and classification of relevant surrounding objects. [48]
- **Interpretation.** Construction and update of faithful representations ("digital twins") of the exogenous operating environment and the endogenous "self" based on perceived inputs and other knowledge sources.
- **Imagination.** Model-based capability of situational awareness, inductive and deductive reasoning, planning, and projection into the future (and the past) based on both exogenous and endogenous world models, derived knowledge, and perceived input.
- **Action.** Selecting and prioritizing appropriate goals for a given configuration of the environmental models, and current beliefs and intents to achieve the selected goals by balancing optimized performance with the need for resilience.
- **Learning.** Ability to adapt and optimize situational behavior; to adapt internal models, goal management, and planning processes dynamically; and to acquire new knowledge through inductive and deductive reasoning.

---

[45] https://www.lexico.com/

[46] The selection of these cognitive capabilities is inspired by "axioms," which are considered necessary for machine intelligence (see also: Aleksander. *Machine Consciousness, Progress in Brain Research*. Elsevier, 2005). The big questions, however, center on cognitive capabilities sufficient for machine consciousness.
[47] See also: Metzler, Shea. *Taxonomy of cognitive functions*. Engineering Design, 2011.
[48] Guidance, Navigation, and Control (GNC)

Situational-aware reasoning, planning, and learning, therefore, are the main ingredients of the *sense–plan–act* cycle (Figure 4), which is the central architectural concept of cognitive actors in the field of *artificial intelligence* (AI).[49] In other words, AI is concerned principally with designing the internals of stream-transforming cognitive actors for mapping from a stream of raw perceptual data to a stream of actions, whereas EI primarily focuses on the perception of and actions on a physical world.

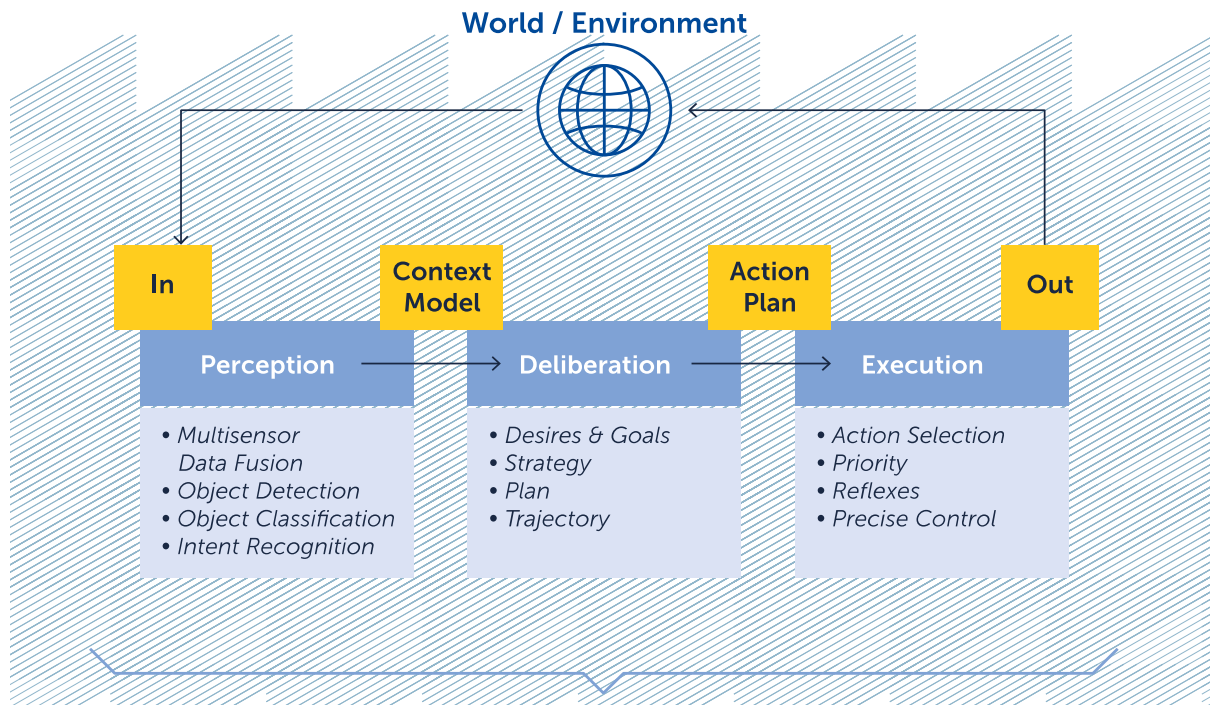

*Figure 4. Sense–plan–act cycle.*

Designs for cognitive actors vary considerably depending on the nature of the operating environment, nature of the perceptual and motor connections between actors and environment, and requirements of the task. For instance, cognitive faculties are categorized as System 1 (fast) capabilities for performing intuitive, automated tasks that we can do instinctively and as System 2 (slow) capabilities for performing tasks that require conscious decision-making and may be described verbally.[50]

There has always been a tendency in AI to use a heterogeneous set of techniques for realizing these two categories of cognitive capabilities, as machine implementations of System 1 functionality are often based on logic and probability theory ("symbolic"), and the implementation of System 2 functionality primarily uses connectionist ("subsymbolic") approaches, which are inspired by the networked neural structure of the brain. Building on these two pillars of AI, *cognitive architectures*, such as Soar[51], integrate logic with connectionist techniques to realize autonomous robotic (both virtual and physical) systems that have some basic cognitive abilities of humans but also in search of a *unified theory of cognition*[52]. With a similar motivation in mind, neurosymbolic programming proposes integrated frameworks that have the pure neural, logical, and probabilistic methods as special cases.[53] Symbolic reasoning capabilities, however, may also be encoded with connectionist networks.

---

[49] Russell, Norvig. *Artificial Intelligence: A Modern Approach*. Pearson, 2021.
[50] According to the global workspace theory of consciousness.

## Intent-driven Systems

Everything an actor needs to know about its goals and expected actions must be defined by means of intents of both human and machine actors. The aim of intent-driven systems is to capture relevant intents and to act in accordance with these intents in an optimized and resilient manner.[54] Intent-driven networks, for example, can predict faults to proactively optimize performance and carry out repairs.[55]

Intents of actors are ideally expressed explicitly and declaratively—that is, as utility-level goals[56] that describe the properties (*what?*) of a satisfactory outcome rather than prescribe a specific solution (*how?*). This gives the system the flexibility to explore various solution options and find an optimized one. It also allows the system to optimize by choosing its own goals that maximize utility. One of the benefits of expressing intents as utility-level goals is that doing so supports the system in coping with conflicting objectives of multiple intents. This is vital because an embodied actor often must take multiple intents into account before making a decision.
Unlike traditional software-intensive systems, where requirements are analyzed offline to detect and resolve conflicts prior to implementation, intents are added to an embodied system and modified during operation. Adaptation to changed intent as well as conflict detection and resolution are, therefore, essential capabilities of embodied systems.

One of the main challenges for specifying goals and intents of embodied systems operating in the real world, however, is that there is very little chance that we can specify our objectives completely and correctly in such a way that the pursuit of those objectives by more capable machines has beneficial outcomes.[57] In particular, we can expect a sufficiently capable machine that is pursuing a fixed objective to take preemptive steps to ensure that the stated objective is achieved, including acquiring physical and computational resources and defending against any possible attempt to interfere with goal achievement. [58]

---

[51] Laird. *The SOAR Cognitive Architecture*. MIT Press, 2012.
[52] Newell. *Unified Theories of Cognition*. Cambridge, 1990.
[53] Raedt, Manhaeve, Dumancic. *Neuro-symbolic = Neural + logic + probabilistic*. IJCAI, 2019.
[54] Silvander, Wnuk, Svahnberg. *Systematic literature review on intent-driven systems*. IET Software, 14(4), 2020.
[55] Huawei. Huawei launches the intent-driven network solution to maximum business value, 2018. (see: http://www.huawei.com/en/press-events/news/2018/2/Huawei-Launches-the-Intent-Driven-Network-Solution)
[56] Including those that may have been considered "common sense" in human-operated systems.

This ability to recognize plans, goals, and intents of other actors enables basic cognitive capabilities to reason about what other actors are doing, why they are doing it, and what they will do next. Consider, for example, a *robotic butler* that does all the boring, repetitive tasks that we wish we did not have to do. Ultimately, we want household assistants that can anticipate our intentions and plans, and provably act in accordance with them. Building these kinds of proactive assistant systems requires a plan, action, and intent recognition for accurately capturing and tracking the operator's requirements.[59]

Plan, goal, and intent recognition also enable different actors to negotiate with each other on behalf of their peers' intent as the basis for a potential mutually beneficial collaboration. For example, one actor may have the intent to deliver high-quality service, while another may want to minimize resource spending. Current AI technology resolves such conflicts either explicitly from weights that introduce relative importance or implicitly from properties of preferential outcomes as defined in utility-level goals.

## Federated Systems

A federated system consists of an ensemble of interacting actors, both machines and humans, that collaborate to jointly perform a common task or achieve a common goal through the mutual provision of actions that individual systems alone cannot achieve.[60] [61]

In doing so, these ensembles exchange relevant information, negotiate their goals, plans, and intentions, and they adapt their own actions to the negotiated plan. This aspect of negotiation—that is, the ability to confer with another as to arrive at the settlement of some matter[62]—is a central activity, and the realization of "confer" and "arrive at settlement" are among the challenges for designing collaborative, federated systems The community of distributed AI in particular has developed a range of models of collaborative plans in the face of resource constraints and uncertainty.

---

[57] *"So tell me what you want, what you really, really want…"* (Spice Girls, Wannabe)
[58] Russell. *Artificial Intelligence and the Problem of Control*. Springer, 2021.
[59] Sukthankar et al (Eds.). *Plan, Activity, and Intent Recognition: Theory and Practice*. Newnes, 2014.
[60] Böhm et al. *Model-based Engineering of Collaborative Embedded Systems*. Springer Nature, 2021.
[61] Grosz. *Collaborative Systems, AAAI-94 Presidential Address*. AI Magazine, 1996.
[62] Mish (Ed.). *Ninth New Collegiate Dictionary*. 1988.

Collaborative federated systems are often safety-critical, operate in dynamic contexts, and must be capable of reacting to unforeseen situations without human inter- vention. On the one hand, they must be able to handle uncertainties due to the imprecision of sensors and the behavior of data-driven components for perceiving and interpreting the context to enable decisions to be made during operation. On the other hand, uncertainties can emerge from the collaboration in a collaborative group related to the exchange of information (e.g., context knowledge) between collaborative systems. Uncertainty that can occur during operation should be considered systematically during engineering to enable collaborating systems to cope with uncertainties autonomously.

Effective collaboration strongly relies on the assumption that most, if not all, actors operate as expected. Therefore, a level of trust and distrust between actors needs to be established as part of the collaboration.[63] Measures need to be taken to tolerate a certain number of non-collaborating and ineffective actors, but also actors with malicious intentions. In particular, the security design of current decentralized systems with their heterogeneous components, including components-off-the-shelf and software of unknown pedigree, start with *a zero-trust* model, which is then mitigated with the right mix of security measures to create challenging barriers for attackers, including pervasive authentication and corresponding checks of all interactions.

## Autonomous Systems

Current industrial practice is mainly concerned with developing remotely operated vehicles ("teleoperated driving") and self-operated systems for restricted time periods and for restricted objectives such as remotely piloted air systems in case of a lost data link, pilotless underwater vehicles, and driverless metros in controlled urban environments. Yet, we are only at the very beginning of a new generation of autonomous systems, which are characterized by *increasingly autonomous* behavior in increasingly complex environments, fulfilling missions of increasing complexity, the ability to collaborate with other machines and humans, and the capability to learn from experience and adapt their behavior appropriately.

As it is designed to perform the equivalent operational tasks of understanding through experiencing and sensing, an autonomous system, therefore, may be viewed as a technical implementation of cognition. These systems perform and integrate heterogeneous sets of tasks on the basis of an overall situational assessment.

In contrast to mere automation, increasingly autonomous systems employ a never-give-up strategy even in the face of real difficulties, such as inconsistencies, unforeseen situations, and authority limits (see "Robotic Companions"). Potential benefits include increased safety, reliability, efficiency, affordability, and previously unattainable mission capability. Clearly, with more autonomy come more and different forms of responsibility.

In the absence of an adequately high level of autonomy that can be relied upon, substantial involvement by human supervisors and operators is required. Increasingly autonomous systems support humans in daily routine tasks, have humans in the loop for continuous control of the evolution of subsystems, and ask humans for high-level decision-making. This kind of mixed human–machine system creates significant new challenges in the areas of human–machine collaboration and mixed-initiative control.

The overarching goal is in achieving a sufficient mutual understanding of the state and intent of both humans and technical systems to optimally blend their competencies in jointly acting toward overarching objectives while respecting privacy. The challenge here is to model human behavior interactions and provide the appropriate uncertainty characteristics related to the largely unpredictable behavior of humans under unforeseen circumstances. Moreover, as individual spheres of control may overlap arbitrarily, these processes need to be orchestrated such that they jointly serve, say, not only a single human but can possibly best multitask in serving arbitrarily large groups at the same time despite uncorrelated requests and uncoordinated missions.

## Self-Learning Systems

One of the distinctive features of embodied systems is their ability to continuously improve their knowledge and capabilities through experience, both positive and negative, and targeted exploration in the real world. This ability of self-learning through exploration is, of course, a far cry from supervised machine learning (ML) schemes for synthesizing, say, neural network representations for approximating functions from given input–output samples. The "learned" behavior may often not be transferred to other operating contexts. An end-to-end autonomous controller, for example, might work well for driving around in the outskirts of Phoenix for which it was trained. However, we most probably expect it to fail miserably when placed on the streets of Algiers or when it is navigating around the Gateway of India.

This is because the neural network controller has not actually learned to drive in the way that we humans are supposed to when visiting a driving school. The controller was just trained to mimic a context-specific set of driving scenarios, and there is no reason to expect it to generalize its behavior to radically new contexts.

In contrast, *autopoietic systems* autonomously extend their perception and attention, situational representation and interpretation of the perceived world, actions, and collaboration patterns, and are able to communicate such learned capabilities with other systems[64]: "*An autopoietic machine is a machine organized (…) as a network of processes of production (transformation and destruction) of components which: (i) through their interactions and transformations continuously regenerate and realize the network of processes (relations) that produced them; and (ii) constitute it (the machine) as a concrete unity in space in which they (the components) exist by specifying the topological domain of its realization as such a network.*" The unsupervised learning ability through interacting with its operating context (including the "self") is the major characteristic of autopoietic systems. This behavior is close to human behavior and possibly also the ultimate dream of AI.

Achieving higher levels of autonomy in open—that is, uncertain, unstructured, and dynamic—environments and terrain increasingly involves data-driven ML techniques with many open-systems science and engineering challenges. The prevalent approach in autonomous driving, for example, aims to reduce the uncertainty of the operating context by compiling and continuously extending large sets of driving scenarios that sufficiently (up to tolerable quantities?) cover all possible situations as the basis for continuous self-learning ecosystems on a global scale.
New and more efficient control regimes for reliable and safe exploration of unknown terrain are clearly needed, as embodied systems must necessarily act with complete, uncertain, and even inconsistent models of the world. Possible approaches that are currently being pursued include the learning and use of causal models, employment of an ensemble of models, and multifaced understanding.[65]

---

[64] Maturana, Varela. *Autopoiesis and Cognition: The Realization of the Living*. Kluwer, 1980.
[65] Minsky: "*You don't really understand something if you only understand it one way.*"

## 3. Trustworthiness

Embodied systems are a new generation of increasingly autonomous systems operating in real-world societal contexts. Thus, the actions of embodied systems do matter.

The autonomous behavior of embodied systems also implies a real danger of losing control because self-learning systems may exhibit emergent behavior, evolve much faster than we as humans may even comprehend, and are able to self-organize in increasingly powerful dynamic federations. These essential features of embodied systems make it even harder to ensure that meaningful (human) control over an embodied system is enabled in the field.

If embodied systems—and the human beings behind them—are not demonstrably worthy of trust, unwanted consequences may ensue, and their uptake might be hindered, preventing the realization of the potentially vast social and economic benefits that they can bring. Trustworthiness, therefore, is a prerequisite for people and societies to develop, deploy, and use embodied systems in a meaningful manner.

Trust may be viewed either as a belief, attitude, intention, or behavior, and as such, it is a complex notion in itself. It is most generally understood as a subjective evaluation of a *trustor* on a *trustee* about something in particular—for example, the completion of a task.[66] A classical definition from organization theory defines trust as the willingness of a party to be vulnerable to the actions of another party on the basis of the expectation that the other will perform a particular action important to the trustor, irrespective of the ability to monitor or control that party.[67] An expert group commissioned by the EC recently identified three main ingredients for trustworthy AI-based systems.[68] Specifically, they recommend trustworthy systems to be at least:

- Lawful. that is, complying with applicable laws and regulations,
- Ethical. that is, ensuring adherence to applicable ethical principles and values,
- Robust. that is, predictable and sustained functionality in the face of uncertainty, faults, and malicious attacks.

Respecting the vast amount of largely unwritten norms for accepted social behavior is also instrumental for building up trust.

Explicit lawful and ethical actors have been proposed based on implementing legal theories, human-like competence, and ethical theories predicated on virtue ethics, deontology, and consequentialism.[69] For example, when ethical principles are in conflict, attempts are made to work out reasonable resolutions. For contexts where informing others of one's intention and reasoning is crucial, these actors communicate and even defend their reasoning.

---

[66] Hardin. *Trust and Trustworthiness*. Russell Sage Foundation, 2002.
[67] Mayer, Davis, Schoorman. *An integrative model of organizational trust*. Academy of Management Review, 1995.
[68] https://digital-strategy.ec.europa.eu/en/policies/expert-group-ai
[69] Scheutz. *The case for explicit ethical agents*. AI Magazine, 2017.

Explicit lawful and ethical actors, however, are inevitably tied to a specific societal context. Therefore, societal-scale CPS that is parameterized by social contexts has been proposed. This approach is based on (1) understanding the nature, scope, and evolution of policies in the operation of societal-scale CPS in different societies, (2) investigating methods for the explicit formal representation of societal context, and (3) developing architectures that guarantee the enforcement of policy requirements.[70]

Table 1. Principles by the EC for responsible and trustworthy artificial intelligence.[71]

| | |
|---|---|
| Human agency and oversight | Including fundamental rights, human agency, and human oversight. |
| Technical robustness and safety | Including resilience to attack and security, fallback plan, and general safety, accuracy, reliability, and reproducibility. |
| Privacy and data governance | Including respect for privacy, quality and integrity of data, and access to data. |
| Transparency | Including traceability, explainability, and communication. |
| Diversity, nondiscrimination, and fairness | Including the avoidance of unfair bias, accessibility and universal design, and stakeholder participation. |
| Societal and environmental wellbeing | Including sustainability and environmental friendliness, social impact, society, and democracy. |
| Accountability | Including auditability, minimization, and reporting of negative impact, tradeoffs, and redress. |

We briefly describe different accounts in the EU and US for trustworthy AI-based systems. Table 1 summarizes the lawful, ethical, and robustness attributes for responsible and trustworthy AI as developed on behalf of the European Commission. Figure 5 displays these practices, which are recommended to be implemented and continuously evaluated throughout the system's life cycle. These are strong recommendations on the use of AI techniques, which seem to be well beyond the current practice for data- and information-driven societal processes.  Notably, most of these recommendations are not specific to AI-ish techniques and may, therefore, be part of any kind of machine ethics. If and how continuous evaluation of these attributes (see Figure 5) can be realized in a practical manner are also unclear, as current auditing frameworks are inefficient, slow, and error-prone for self-learning and ever-evolving systems, and we currently do not have adequate technical means for automated compliance audits. Moreover, tradeoffs usually need to be made in real-life engineering to address seemingly contradictory attributes such as privacy and transparency.

---

[70] NSF PIRE 16-571: *Science of Design for Societal-Scale Cyber–Physical Systems*
[71] https://www.aepd.es/sites/default/files/2019-12/ai-ethics-guidelines.pdf

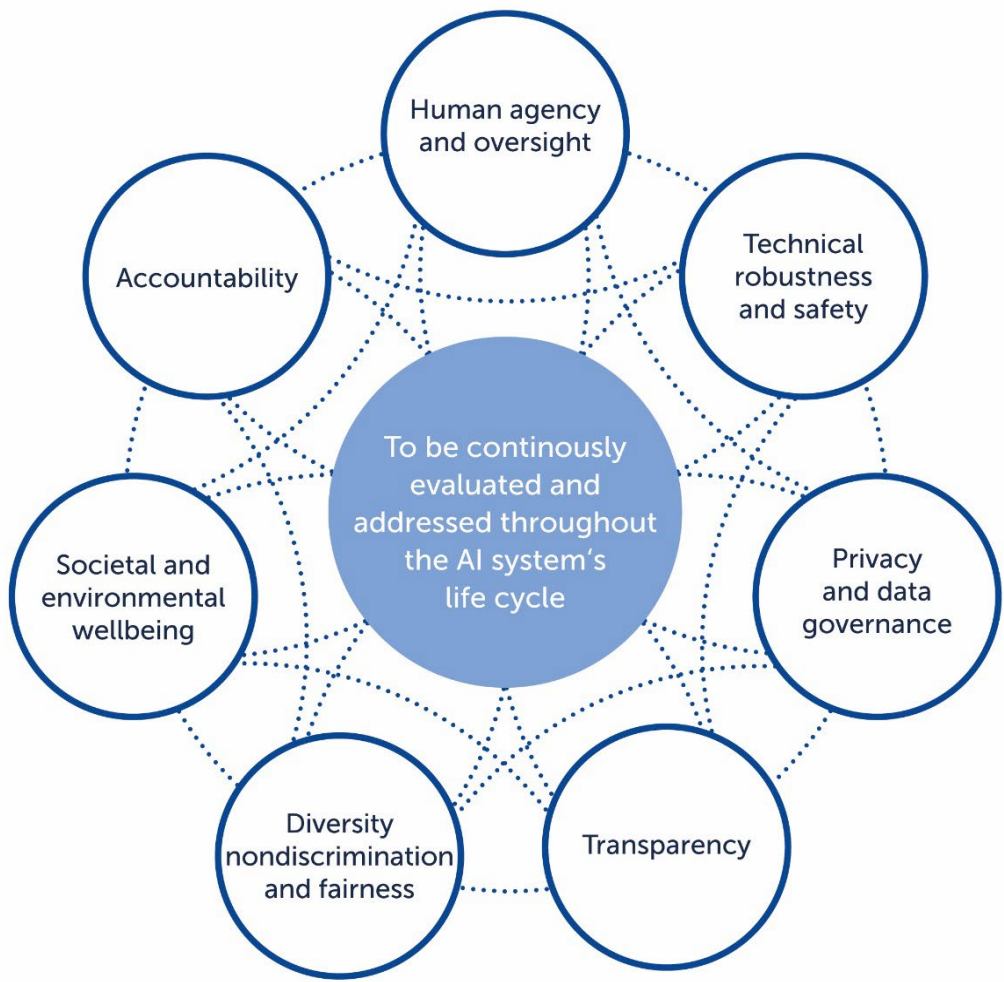

*Figure 5. Attributes of trustworthy artificial intelligence.[72]*

---

[72] Source: EU High-Level Expert Group on AI (https://www.aepd.es/sites/default/files/2019-12/ai-ethics-guidelines.pdf)

In a similar vein, the principles in Table 2 by the ACM for transparency and accountability [73] are designed to increase trust in all kinds of algorithmic systems.

Table 2. ACM principles for transparency and accountability.[74]

| | |
|---|---|
| Awareness | Owners, designers, builders, users, and other stakeholders of analytic systems should be aware of the possible biases involved in their design, implementation, and use and the potential harm that biases can cause to individuals and society. |
| Access and redress | Regulators should encourage the adoption of mechanisms that enable questioning and redress for individuals and groups that are adversely affected by algorithmically informed decisions. |
| Accountability | Institutions should be held responsible for decisions made by the algorithms that they use, even if explaining in detail how the algorithms produce their results is not feasible. |
| Explanation | Systems and institutions that use algorithmic decision-making are encouraged to produce explanations regarding both the procedures followed by the algorithm and the specific decisions that are made. This is particularly important in public policy contexts. |
| Data provenance | A description of the way the training data was collected should be maintained by the builders of the algorithms, accompanied by an exploration of the potential biases induced by the human or algorithmic data-gathering process. Public scrutiny of the data provides maximum opportunity for corrections. However, concerns over privacy, protecting trade secrets, or revelation of analytics that might allow malicious actors to game the system can justify restricting access to qualified and authorized individuals. |
| Auditability | Models, algorithms, data, and decisions should be recorded so that they can be audited in cases where harm is suspected. |
| Validation and testing | Institutions should use rigorous methods to validate their models and document those methods and their results. In particular, they should routinely perform tests to assess and determine whether the model generates discriminatory harm. Institutions are encouraged to make the results of such tests public. |

---

[73] ACM Code of Ethics (http://ethics.acm.org)
[74] *ACM Code of Ethics (http://ethics.acm.org)*

We may observe a strong overlap of the trustworthiness requirements in the EU and US because of the relative similarity of the underlying systems of social values. However, the ACM principles for transparency and accountability tend to be phrased more in technical terms; consequently, they seem to be more amenable to automated compliance. The ACM principles for transparency and accountability were clearly formulated with current AI/ML techniques in mind. As such, they do not adequately address the characteristics of the upcoming generation of increasingly autonomous and self-learning systems. Particularly, clearly identifiable "institutions" anymore for operating an embodied system may no longer exist, and embodied systems might eventually need to be held responsible for their very own actions.

Moreover, the demand for data provenance as formulated above may not be applicable to continuous, unsupervised learning. Depending on the intended role of embodied systems in societal contexts, ranging from mere assistants to fully autonomous actors, the ACM principles for transparency and accountability, therefore, need to be revised, developed further, and agreed upon to fit the characteristics of a new generation of increasingly autonomous and self-learning systems. We also need to develop techniques for ensuring algorithmic transparency and accountability for dynamic federations of increasingly autonomous, learning-enabled, and embodied systems.

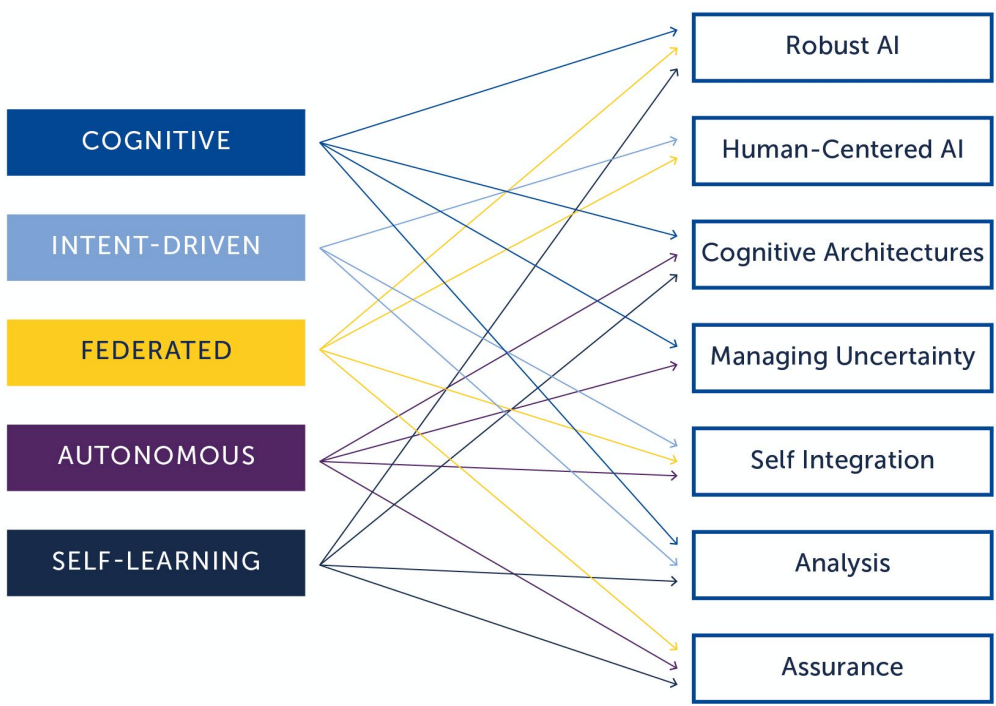

*Figure 6*. Mapping from main characteristics of embodied systems to corresponding system challenges. AI, artificial intelligence.

# 4. Challenges

Traditional system engineering comes to a juncture from assuring the quality of service, dependability, and safety attributes for relatively small-scale, centralized, deterministic, and predictable, nonevolvable, automated *embedded systems* operating in well-defined and predictable environments to assure the trustworthiness of larger-scale, federated, unpredictable, self-learning, and increasingly autonomous *embodied systems* operating in uncertain and largely unknown environments. These differences between embedded and embodied systems are summarized in Figure 7.

|  | Embedded systems | Embodied systems |
| --- | --- | --- |
| Architecture | Centralized | Federated |
| Behavior | Deterministic | Largely unpredictable |
| Context | Well defined | Uncertain |
| Maintenance | Updates | Self-learning |
| Requirement | Dependability | Trustworthiness |
| Human control | Yes | Increasingly no |

*Figure 7. From embedded to embodied systems.*

Thereby, we can build up trust by engineering embodied systems that are lawful, ethical, and robust. Compared with traditional *embedded systems engineering*, we face the following additional challenges, particularly in embodied systems:

- Learn continuously and adapt and optimize their behavior based on experience and targeted exploration.
- Need to safely operate in partially unknown or uncertain environments, and they need to be robust in the presence of inaccuracies, uncertainty, and errors in their world models ("known unknown") and also in the presence of nonmodeled phenomena ("unknown unknown").
- Increasingly lack fallback to a responsible human being.

- Offer a variety of new attack surfaces due to data-driven programming.[75]
- Exhibit largely unpredictable and emergent behavior due to data-driven programming.
- Cannot be certified because current certification regimes require the system's behavior and its intended operating context to be fully specified and verified prior to commissioning.

For the continuous evolvability and self-organizing capabilities of embodied systems, the traditional *design–build–commission–decommission* life cycle of embedded systems is inadequate and clearly needs to be replaced with a *design–operation continuum* based on the combined functionality for design, simulation/verification, deployment, operation, and maintenance.

Given the expected lifetime of embodied systems, they must be designed to cope with changing underlying technologies and hardware, regulatory settings, societal contexts, system requirements, and yet provide the same high level of dependability and quality of service throughout their evolution. It is even perceivable that certain design and engineering steps, including situational risk management, are eventually performed by the embodied systems themselves.

On the basis of the characteristics of embodied systems as outlined in Section 2, we are now deriving all-important systems challenges for developing, deploying, and operating trustworthy embodied systems. A high-level summary of this derivation is provided in Figure 6. However, this figure depicts the most obvious and possibly the most important relations between characteristics and derived challenges for engineering trustworthy embodied systems.

## Robust Artificial Intelligence/Machine Learning

ML techniques are ubiquitous and have many successful applications. The main attraction of ML is that functional requirements are stated in terms of data only, and a corresponding program for approximating such a function is synthesized in an automated fashion. Moreover, engineering steps from data wrangling to architectural selection and optimization of the function approximation are increasingly being automated.[76] However, these advantages also come with some downsides.

- Current learning techniques based on artificial neural networks (ANNs), for example, are only as good as the available data (and their resource-intensive preprocessing and labeling requirements).

---

[75] vulgo: machine learning.
[76] Xin, Zhao, Chu. AutoML: A survey of the state-of-the-art. *Knowledge-Based Systems*, 212, 2021.

- Uncertainty exists with regard to the input–output behavior of the learned function approximators usually because they are not robust in the sense that small changes to the input might result in unexpected behavior; for example, one-pixel attacks of trained classifiers are successful for many ANNs.
- The learned function is usually restricted to the context as encoded in the learning data with limited transferability.[77]
- The input–output behavior of the learned function is implicit and not transparent in that the reasons for proposed decisions are opaque. For example, a neural network for classifying tumors as benign or not might show human-like performance, but further explanation is usually given to the human diagnostician.
- Current supervised learning algorithms based on stochastic gradient descent, for example, are rather data-intensive and inefficient. Small children, in particular, have the ability to form concepts such as "cow" on the basis of only a few encounters.

As we gain more experience with developing and deploying ML-based systems for real-world challenges, we realize some shortcomings.[78]

- Adequate data often are not or not readily available, and crucial data for specifying the intended behavior might become available during development and operation only. Traditional process models with clearly defined phases of requirements engineering, however, do not adequately support the extra flexibility needed by a data-driven AI development approach.
- Building data-driven ML systems requires a comprehensive wealth of experience, even though this kind of expert knowledge ("Which learning algorithm?," "Which architecture?," "Which hyperparameters?") is increasingly being captured in automated metalearning processes.[79]
- An ML system is often more than just ML; building and running it is a serious software and systems engineering undertaking. However, generally accepted or even standardized processes, methods, and tools for the development and operation of predictable and transparent ML are largely missing.

---

[77] *"The soccer bot lines up to take a shot at the goal. But instead of getting ready to block it, the goalkeeper drops to the ground and wiggles its legs. Confused, the striker does a weird little sideways dance, stamping its feet and waving one arm, and then falls over. 1-0 to the goalie… AI trained using reinforcement learning can be tricked by … an adversarial policy."* (quoted from Heaven. *Reinforcement learning AI are vulnerable to a new kind of attack*. MIT Technology Review, Artificial Intelligence, 2020)
[78] Standards for engineering trustworthy autonomous systems are currently emerging, in particular, VDE-AR-E 2842-61 of the standardization organization DKE (cmp. https://www.dke.de/de/normen-standards/dokument?id=7141809&type=dke|dokument )
[79] For example, AutoML.

- Current ML applications usually do not adequately address trustworthiness attributes, as outlined in Section 3.

Moreover, most ML applications nowadays are based on supervised learning, which does not support the required self-learning capabilities of embodied systems. However, reinforcement learning, which is based on optimizing objective functions, is a good starting point, where inverse reinforcement learning might be used for intent recognition. However, current approaches to reinforcement learning require a large number of trials. Failed attempts in the real world might result in undesired behavior, accidents, or other catastrophic events.

Techniques for safe and predictable self-learning and exploration of uncertain and unpredictable real worlds are still in their infancy. Promising approaches for tackling the added complexities of autonomous actors in real-world settings include hybrid combinations[80] of classical and learning-based algorithms and the incorporation of prior knowledge.[81]

How can one learner who does not know what there is to learn, manage to learn anymore? Current ML approaches usually start with what needs to be learned for learning. We, as humans, however, can discover both the tasks to be learned and the solution to those tasks through exploration or nongoal-directed action.

ML has mainly concentrated on nonincremental learning tasks, where the entire training set is fixed at the start of learning and then is either presented in its entirety or randomly sampled. Embodied actors, however, need to learn incrementally and continuously through exploration. ML is also increasingly being augmented with domain-specific knowledge, and rules to increase the efficiency and effectiveness of ML, rules, and decision trees might also be compiled from learned behavior, which themselves can be used to improve further learning and make decisions transparent, say, to a human operator. In this way, domain knowledge, such as physical laws, is currently integrated into ML using techniques such as regularization, data augmentation, or postprocessing.

A recent survey on knowledge-augmented ML[82] reviews the role of knowledge in ML and discusses its relation to the concept of invariance. *Neurosymbolic integration* (with logic, probability theory, and neural structures as projections) has been proposed as the basis for a new generation of dependable, predictable, transparent, and efficient data-driven programming techniques for realizing increasingly autonomous human assistants and/or for mission- and safety-relevant applications.

---

[80] Chaplol et al. *Learning to explore using active neural slam*. ICLR, 2020.
[81] Ye, Yang. *From seeing to moving: As survey on learning for visual indoor navigation*. arXiv:2002.11310, 2020.
[82] https://www.fortiss.org/fileadmin/user_upload/05_Veroeffentlichungen/Whitepaper/fortiss_whitepaper_knowledge_as_invariance_web.pdf

Despite technological advances that have led to the proliferation of data-driven ML systems, the question of the level of trust that we can put in these systems remains. What is needed, therefore, is a new generation of *robust ML* algorithms that,[83]

- can operate in uncertain and largely unpredictable environments,
- can make timely and confident decisions,
- produce results that are understandable and explainable to a human operator,
- are resilient to erroneous inputs and targeted attacks,
- can process ever-increasing amounts of data,
- come from decentralized and heterogeneous data sources,
- can extract useful insights from small amounts of data and sparse rewards,
- do not make significant compromises in confidentiality and privacy in federated multiactor settings.

Traditionally, ML modeling techniques have relied on *unsiloing data* from multiple sources into a single *data lake*. Centralized data sources, however, pose serious privacy, data misuse, and security challenges for federated systems. Moreover, aggregating diverse data from multiple sources need to meet regulatory principles such as the General Data Protection Rights (GDPR), the Health Insurance Portability and Accountability Act (HIPAA), or CCPA (we will reconsider these issues in the subsection below on Assurance). To overcome these challenges, several pillars of *privacy-preserving ML* have been developed for *unsiloing ML models* with specific techniques that reduce privacy risk and ensure that data remain reasonably secure, namely, federated ML, secure multiparty computation, differential privacy, and homomorphic encryption. Altogether, the prevailing methods for ML do not map to the ways that humans learn, as humans learn by seeing, moving, interacting, and speaking with others. Humans learn from sequential experiences, not from shuffled and randomized experiences. We need to come up with a new generation of ML techniques that possibly mimic the ways humans learn to enable efficient self-learning for trustworthy embodied systems through targeted exploration and experience.

---

[83] Stoica et al. *A Berkeley View of Systems Challenges for AI*. 2017 (https://arxiv.org/pdf/1712.05855.pdf).

## Human-Centered Artificial Intelligence/Machine Learning

Embodied systems are machines that support humans in daily routine tasks, have humans in the loop for the continuous supervision of the evolution of subsystems, and ask humans to undertake high-level decision-making. The central challenge, as addressed in the field of human-centered engineering, is to enable symbiotic relationships, in which embodied systems and humans augment each other reciprocatively,[84] and as the basis for coevolutionary improvement of both machines and humans.[85]

Humans and machines need to avoid "mode confusions" based on a mutual understanding of state and intent of both humans and machines to optimally blend their competencies in jointly acting toward overarching objectives while respecting privacy. Moreover, in the absence of an adequately high level of autonomy that can be relied upon, substantial involvement by human supervisors and operators is required. This need creates significant new challenges in the areas of human–machine interaction and mixed-initiative control.

Embodied systems are learning enabled. However, as discussed above, current ML techniques offer new attack surfaces, are largely nontransparent (implicit models), tend to be energy and data-hungry, and lack basic transferability capabilities that are required to navigate unknown or uncertain terrain.[86] Therefore, if and how these technologies can be used beneficially in real-world applications that require human–machine interaction or in mission- and safety-critical applications is unclear.[87]

Moreover, the apparent success of ML in producing seemingly intelligent decisions poses the risk of misunderstanding during the communication between humans and machines. A comparison of the behavior of ML systems and humans in decision-making shows obvious significant differences. ML essentially provides efficient algorithmic solutions for optimizing a well-defined target function, enabling the learning of task- and data-specific patterns from a large number of samples or observations.

---

[84] Cmp. IBM high-level AI framework.
[85] Engelbart. *The Bootstrap Paradigm* (https://dougengelbart.org/content/view/248/).
[86] For example, the keynote of AAAI President, http://web.engr.oregonstate.edu/~tgd/talks/dietterich-aaai-presidents-address-final.pdf.
[87] See https://www.fortiss.org/fileadmin/user_upload/05_Veroeffentlichungen/Whitepaper/fortiss_whitepaper_HCML_web.pdf.

In contrast, a human would rather make decisions based on ground truth rules, such as causality, and can transfer known solutions to new situations and domains. Although both types of decision-making can be called forms of generalization, the human way of decision-making is a harder form of generalization, sometimes termed horizontal, strong, or out-of-distribution generalization. Human decision-making takes advantage of heterogeneous information sources, such as interventions, domain shifts, and temporal structures, which ML typically discards or even fails to model during learning processes.

These shortcomings of ML lead to serious challenges in designing trustworthy systems based on ML for human users, such as:

- **Low explainability.** The decision-making mechanism of ML algorithms, such as ANNs, cannot be made fully transparent to humans and is difficult for humans to interpret.
- **Miscalibration of trust.** ML seems to be both highly effective to humans and also largely predictable, thereby luring humans into accepting these technical systems as human-like partners (anthropomorphization) that are trusted more than actually justified.[88]
- **Low level of human control and involvement.** Most ML algorithms rely on either a hypothetical model of the distribution of data or concrete interpretation (labeling) of data. Such constructions have become one major hurdle to enabling ML systems with high levels of human control, such as human-like reasoning and generalization. Specifically, finding the right level of human control on which the system can effectively communicate with humans to obtain such input is difficult.

The challenge here is to model human behavioral interactions with the technical system and provide the appropriate uncertainty characteristics related to the largely unpredictable behavior of humans under unforeseen circumstances. Moreover, as individual spheres of control may overlap arbitrarily, these processes need to be orchestrated such that they not only jointly serve a single human but can also possibly best multitask in serving arbitrarily large groups at the same time despite uncorrelated requests and uncoordinated missions.
This is particularly challenging for complex mission tasks that call for collaboration and teamwork among humans and machines. In these cases, AI-enabled systems may need to identify the real intent of human operators and their goals, interact with them in a goal-oriented manner based on models of human behavior, and, in extreme cases, tolerate and adequately mitigate seemingly irrational behavior.

---

[88] This phenomenon was previously demonstrated by Feigenbaum's ELIZA program.

Building trust based on explainability ("how?," "why?," "what if?") is essential for human operators to accept ML-based solutions and those systems that incorporate decisions made by them.

Explainability, therefore, is particularly useful for

- increasing the confidence of human operators,
- building trust by supporting an increased understanding of the transferability of results to other problems of interest,
- avoiding misconception and ensuring that humans understand outcomes of learning-enabled components as a solid basis for intervening human actions, and
- increasing human confidence in the decisions and predictions made by a learning-enabled component.

However, significant challenges remain in developing adequate methods for explainability. One of them is the tradeoff between attaining the simplicity of algorithmic transparency and impacting the high-performing nature of complex but opaque ML models. Yet another challenge is to identify the right information for the user, where different levels of knowledge will come into play. Beyond selecting the level of knowledge retained by the user, generating a concise explanation also becomes a challenge. Most existing methods for explainability, however, focus on explaining the processes behind an ML-based decision, which is often useless in a particular application domain. In addressing these issues, current research is integrating ML algorithms with domain-specific (for example, laws of physics) and, possibly learned, knowledge.

## Cognitive Architectures

The field of cognitive architectures creates programs that can reason about problems across different domains, develop insights, adapt to new situations, and even reflect on themselves. These programs realize cognitive functions, including perception, memory, attention, social interaction, planning, motivation, actuation, reasoning, communication, learning, emotion, modeling self/other, building/creation, and arithmetic capabilities.

Prominent cognitive architectures include Soar, ACT-R, LIDA, CLARION, and EPIC.[89] However, with no clear definition and the general theory of cognition, several hundreds more architecture based on different sets of premises and assumptions are available, also coming from various backgrounds (computer science, psychology, philosophy, and neuroscience).[90]

---

[89] Nancy, Balamurugan, Vijakkumar. *A comparative analysis of cognitive architecture.* JARTET, 2016.

What constitutes a cognitive architecture is not even clear at all. Newell's criteria for a cognitive architecture, for instance, include flexible behavior, real-time behavior, rationality, large knowledge base, learning, development, linguistic abilities, self-awareness, and brain realization.[91] Sun's desiderata are broader and include, among others, cognitive realism, adaptation, modularity, routineness, and synergistic interaction.[92] Many of these criteria for cognitive systems are clearly of general interest also for the class of embodied systems, as defined above.

The obvious question, therefore, is if and how principles of cognitive architectures are aiding in the design of embodied systems.[93] For example, machinery that is expected to behave "correctly" in a complex world is recognized to possibly be akin to a model-based reflective predictive controller of a machine with a mission.[94] [95]

Some cognitive architectures use one uniform representation and corresponding learning method that yields "grand unification and functional elegance"[96] but loses expressiveness. Others utilize general knowledge representations and many inference strategies[97] that result in higher expressiveness, but they cause difficulties with integrations of different components of the cognitive architecture. A substantial number of cognitive architectures are hybrid (for example, Soar, ACT-R, LIDA, CLARION, EPIC) in that they combine symbolic and subsymbolic reasoning, thereby providing architectural concepts for integrating connectionist[98] with logic-based AI technologies.

Probabilistic programming provides yet another framework in which basic components of cognitive architectures are represented in *a unified and elegant fashion*.[99] This probabilistic model cognition is destined to support aleatoric uncertainty, that is, the "known unknown." Probabilistic programming also suggests a programming model for embodied systems based on well-known concepts of program construction in computer science for specifying, developing, analyzing, synthesizing, and composing programs.

---

[90] Kotseruba et al. *A review of 40 years of cognitive architecture research*. arXiv:1610.08602, 2016.
[91] Anderson, Lebiere. *The Newell test for a theory of cognition*. Behavioral and Brain Sciences, 26(5), 2003.
[92] Sun. *Desiderata for cognitive architectures*. Philosophical Psychology, 17(3), 2004.
[93] Irrespective of their relative ability to explain the development of higher-level intelligent behavior and consciousness.
[94] Sanz, Lopez, Rodriguez, Hernandez. *Principles for consciousness in integrated cognitive control*. Neural Networks Society, 20(9), 2007.
[95] Sanz. *Thinking with the body: Towards hierarchical scalable cognition*. Handbook of Cognitive Science, An Embodied Approach. Elsevier, 2008.
[96] Rosenbloom. *Extending Mental Imagery in Sigma*, Springer, 2012.
[97] Goertzel, Pennachin, Geisweiller. *Engineering General Intelligence*. Atlantis Press, 2014.
[98] Connectionist architectures are supposed to exhibit intelligent behavior without storing, retrieving, or otherwise operating on structured symbolic expressions.
[99] Potapov. *A Step from Probabilistic Programming to Cognitive Architectures*. arXiv, 2016.

Numerous demonstrations of cognitive architectures for performing real-world tasks are available, including navigation, obstacle avoidance, object manipulation, fetch-and-carry tasks for trash collecting,[100] and soda-collecting[101] mobile robots. Applications from industrial domains include robotic crane operation,[102] bridge construction,[103] autonomous cleaning and deburring workstation,[104] an automated stamp distribution center,[105] and an analytics engine, as inspired by the HTM cognitive architecture.[106] Cognitive architectures have also proven to be useful for human performance modeling, human–robot interaction, natural language processing, categorization and clustering, and computer vision.

Cognitive architectures might be able to support *active perception*[107] for coupling perception with the action of an embodied actor. For example, an actor may be spawned anywhere in the environment and may not immediately "see" the pixels that contain the answer to its visual goal (for example, the car/goal may not be visible). Thus, the actor must move to succeed, controlling the pixels that it will perceive. The actor must learn to map its visual input to the correct action on the basis of its perception of the world, the underlying physical constraints, and its understanding of the question. The observations that the actor collects are a consequence of the actions that the actor takes in the environment, and the actor is controlling the data distribution that is coming in. The actor controls the pixels it gets to see. One of the challenges of active perception is to be generally robust to variation.

Finally, cognitive architectures and theories from psychology, such as *cue theory*, might serve as the basis and inspiration for designing novel control regimes for embodied actors capable of safely exploring and navigating the "unknown unknown." In this way, careful terrain exploration has been approached by minimizing surprises, for example, based on active inference[108] and the free energy principle[109] [110] or, alternatively, by maximizing predictive information.[111]

---

[100] Firby et al. *An Architecture for Vision and Action*. IJCAI, 1995.
[101] Brooks. *A robot that walks: Emergent behaviors from a carefully evolved neural network*. Neural Computation, 1(2), 1989.
[102] Lytle, Saidi. *NIST research in autonomous construction*. Autonomous Robots, 22(3), 2007.
[103] Bostelman, Bunch. *Delivery of an advanced double-hull ship welding*. ICSC, Symposia on Intelligent Industrial Automation and Soft Computing, 1999.
[104] Murphy, Norcross, Proctor. *CAD directed robotic deburring*. Robotics and Manufacturing Research, Education, and Applications, 1988.
[105] Albus. *The NIST real-time control system (RCS): An approach to intelligent systems research*. Journal of Experimental & Theoretical Artificial Intelligence, 9(2–3), 1997
[106] https://numenta.com/grok
[107] Aloimonos (Ed.). *Active Perception*. Psychology Press, 1993.
[108] Active inference: maintaining a model and its predictions through action to change the sensory inputs to minimize prediction error indirectly (if the sound is not getting louder, then one should move closer to the train to hear the train getting louder).
[109] Friston's free energy principle (FEP) is a leading formal theory of self-organizing system dynamics. It basically asserts that living systems must minimize the entropy of their sensory exchanges with the world; for example, Friston. *The free-energy principle: A unified brain theory?*, Nature Reviews Neuroscience, 11(2), 2010).
[110] Smith. *A Unified Framework for Intelligence based on the Free Energy Principle*. 2019.
[111] Ay, Bertschinger, Der, Güttler, Olbrich. *Predictive information and explorative behavior of autonomous robots*, European Physical Journal B, 63(3), 2008.

## Uncertainty Quantification

A multitude of sources is available for uncertainty in the design and operation of embodied systems, given the controller uncertainty due to nondeterminism, probabilistic control algorithms, and the uncertainty about the following: their operational context (for example, how many and which objects and actors are in the environment), corresponding hazards and risks, the behavior of learning-enabled components, safety envelopes, the internal models, and, last but not least, the intentions, behaviors, and strategies and actions of other embodied actors, both human[112] and machines.

Learning in the sense of replacing specific observations using general models is a process of inductive inference. Such models are never provably correct but only hypothetical and, therefore, uncertain; the same holds true for the predictions produced by a model. For example, the input–output behavior of ANNs heavily relies on the selection of "complete" and "correct" sets of training and support data for faithfully specifying relevant operating contexts (input) and their intended internal representation (output). Another source of uncertainty for ANNs is the use of stochastic search heuristics, which may lead to incorrect recall even for inputs from the training data, and the largely unpredictable capability of generalizing from given data points. Uncertainty on the faithfulness of the training data that represent operating contexts and uncertainty on the correctness and generalizability of training also combine in a, well, uncertain manner.

One usually distinguishes between *aleatoric* and *epistemic* sources of uncertainty. Aleatoric[113] uncertainty refers to the variability in the outcome of an experiment due to inherently random effects, and epistemic[114] uncertainty refers to uncertainty caused by a lack of knowledge.[115] For example, incomplete knowledge of an embodied actor's operating context is an epistemic source of uncertainty. As epistemic uncertainty refers to the ignorance of an actor and hence to its epistemic state, it can, in principle, be reduced with additional information.

---

[112] Compare with the Human-Centered Artificial Intelligence/Machine Learning challenges above.
[113] Also known as statistical, experimental, or "known unknown."
[114] Also known as systematic, structural, or "unknown unknown."
[115] Hüllermeier, Waegeman. *Aleatoric and epistemic uncertainty in machine learning: An introduction to concepts and methods.* Machine Learning, 110(3), 2021.

The central challenge is *uncertainty quantification*,[116] that is, to systematically reduce uncertainty to an acceptable level and serves as the basis for trustworthy and (up to tolerable quantities) predictable embodied systems. Uncertainty quantification involves:

- Identifying all relevant sources of uncertainty.
- Adequately quantifying and estimating uncertainty.
- Understanding how uncertainty accumulates, forward and inverse, along the chains of computations.
- Reducing overall uncertainty below acceptable levels.[117]
- Managing incremental change of uncertainty.

Table 3 lists corresponding challenges for the rigorous design of embodied systems and for managing uncertainties throughout its life cycle. The analysis and assurance challenges are also addressed in the corresponding subsections on Analysis and Assurance.

Different uncertainty-reducing techniques are available for robust AI systems depending on the aleatoric or epistemic nature.[118] The basic principle of uncertainty reduction also plays a key role in active learning[119] and in learning algorithms.[120] For example, indirect *cues*[121] may cause the system to hypothesize the existence of a relevant object, which needs to be confirmed using additional actions. Additionally, *uncertainty quantification* approaches in engineering have been designed to demonstrate that, with high probability, a real-valued response function of a given physical system does not exceed a given safety threshold.[122]

---

[116] Uncertainty quantification (UQ) is the science of quantitative characterization and reduction of uncertainties in both computational and real-world applications. It tries to determine how likely certain outcomes are if some aspects of the system are not exactly known.
[117] For example, less than one hazardous behavior for $10^9$ operational times.
[118] Dietterich. *Steps toward robust artificial intelligence*. AI Magazine, 38(3), 2017.
[119] Aggarwal, Kong, Gu, Han, Philip. *Active learning: A survey*. Data Classification: Algorithms and Applications. CRC Press, 2014.
[120] Mitchell. *The need for biases in learning generalizations*. Tech. Rep. TR CBM–TR–117, Rutgers University, 1980.
[121] Björkman. *Internal cue theory: Calibration and resolution of confidence in general knowledge*. Organizational Behavior and Human Decision Processes, 58(3), 1994.
[122] Owhadi, Scovel, Sullivan, McKerns, Ortiz. *Optimal uncertainty quantification*. Siam Review, 55(2), 2013.

Table 3. Engineering challenges for uncertainty quantification.

| | |
|---|---|
| Specification challenge | • Provide means for constructing (and maintaining) safety envelopes either deductively from safety analysis or inductively from safe nominal behavior.<br>• Provide means for minimizing uncertainties related to safety envelopes with a given level of effort.<br>• Provide means for deriving safety requirements for learning-enabled components, which are sufficient for establishing artificial intelligence (AI) system safety.<br>• Provide means for reducing specification uncertainty by means of deriving data requirements for learning-enabled components. |
| Prediction challenge | • Identify all relevant sources of uncertainty for an AI system.<br>• Provide adequate means for measuring uncertainty.<br>• Calculate forward propagation of uncertainty, where the various sources of uncertainty are propagated through the model to predict overall uncertainty in the system response.<br>• Identify and solve relevant inverse[123] uncertainty quantification problems for safe AI (using, for example, a Bayesian approach).<br>• Predict (up to tolerable quantities) the unsafe behavior of AI systems operating in uncertain environments. |
| Assurance challenge | • Provide adequate measures of uncertainty to assure AI system safety.<br>• Construct and maintain evidence-based arguments to support the certainty and rebut the uncertainty of safety claims.<br>• Identify useful safety case patterns for safe AI systems together with compositional operators on safety cases for managing uncertainty. |
| Design challenge | • Develop safety case patterns for different architectural designs of AI systems.[124]<br>• Construct safe and quasipredictable AI systems compositionally together with their safety cases. |
| Analysis challenge | • Provide adequate means for measuring and reducing uncertainty on the input–output behavior of learning-enabled components.<br>• Define and measure the respective contributions of static and dynamic analysis techniques for learning-enabled systems to reduce safety-related uncertainty to tolerable levels. |
| Maintenance challenge | • Identify incremental change operators for maintaining uncertainty and safety assurance of self-learning AI systems.<br>• Safely adapt and optimize the situational behavior of an AI system (together with its safety cases based on the principle of minimizing uncertainty). |

---

[123] That is, calculating from a set of observations the causal factors that produced them.
[124] In analogy to, say, Mils separation kernel protection profile.

Uncertainty quantification also plays a pivotal role in reducing uncertainties for learning-enabled components such as ANNs.[125] [126] Establishing resilience[127] and other invariance[128] properties, for example, is an important means to reduce the behavioral uncertainty of ANNs. Moreover, measuring and estimating the uncertainty of the input–output behavior of learning-enabled components is essential for, say, switching between performant and safe channels in a simplex architecture, and uncertainty information is useful input for planning safe actions.

Proposals for measuring behavioral uncertainty of learning-enabled components include the following:

- The distance between neuron activations observed during training and the activation pattern for the current input are used to estimate the input–output uncertainty.[129]
- Ensemble learning techniques are used to estimate input–output uncertainty by training a certain number of ML components from different initializations and sometimes on differing versions of the dataset; the variance of the ensemble's predictions is then interpreted as to its epistemic uncertainty.
- Certain instances of ensemble learning techniques, such as Bayesian neural networks, measure both epistemic uncertainties on model parameters and the aleatoric uncertainty of the input–output behavior with respect to model parameters.[130]

What we should focus on, however, is not so much about reducing behavioral uncertainty of individual components but of the embodied system itself. Such uncertainty on the system-level behavior is obtained, for example, by forward propagation[131] of component uncertainties along chains of computation.

Uncertainties can also be explicitly managed through *assurance cases*.[132] These structured arguments are comprehensive, defensible, and valid justification that the system fulfills crucial properties, at least up to a tolerable level of uncertainty, with the goal of increasing confidence and building up trust in the behavior of an embodied system. The purpose is, broadly, to

---

[125] Czarnecki, Salay. *Towards a framework for managing perception uncertainty for safe automated driving*. Computer Safety, Reliability and Security, 2018.
[126] Abdar et al. *A review of uncertainty quantification in deep learning: Techniques, applications and challenges*. Information Fusion, 76, 2020.
[127] Cheng, Nührenberg, Rueß. *Maximum resilience of artificial neural networks*. International Symposium on Automated Technology for Verification and Analysis, 2017.
[128] With respect to certain classes of input transformations such as stretching.
[129] Cheng, Nührenberg, Yasuoka. Runtime monitoring neuron activation patterns. Design, Automation & Test in Europe Conference & Exhibition, 2019.
[130] Jospin, Buntine, Boussaid, Laga, Bennamoun. *Hands-on bayesian neural networks-a tutorial for deep learning users*. *arXiv:2007.06823*, 2020.
[131] For example, based on Bayesian inference.
[132] We will be revisiting assurance cases in the subsection describing the Assurance challenge.

demonstrate that the crucial risks associated with specific system concerns[133] have been identified, are well understood, and have been appropriately mitigated, and that there are mechanisms in place to monitor the effectiveness of defined mitigations. Of particular interest is to see how the influence of a learning-enabled component is captured and reasoned within the control structure of an embodied system. Recent extensions of assurance cases for reasoning about confidence and uncertainty seem to be a good starting point for a more thorough investigation into uncertainty quantification for embodied systems. [134] [135]

Altogether, interest in various aspects of uncertainty quantification for embodied systems has been increasing. What is still missing, however, is a comprehensive set of methods and tools for the rigorous design of embodied systems based on the principle of uncertainty quantification.

## Self-Integration

Let's reconsider the federated retail example in Figure 3. How and why do all the embodied actors, ranging from design, production, and logistics, form a collaborative federation in a productive manner, thereby supporting the intent of the buyer? Moreover, how does this federation tolerate real-world mishaps, such as a ship getting stuck in a channel?
Intent-driven formation of purposeful federations of embodied systems requires the individual systems to be *open* to collaborating with others while still operating as self-sufficient individually purposeful systems. The formation of these federations is based on *self-integration*, which seeks out other systems to support to meet their local and global intents and goals, which cannot be accomplished on their own.
The Semantic Interoperability Logical Framework (SILF)[136], for example, facilitates dependable machine-to-machine information exchange based on an extensive ontology to describe the content of messages and an intent-aware mediation mechanism to translate messages as needed. These adapters may be synthesized automatically from ontological descriptions, whereas the purpose of the integration is represented in a task ontology.[137] SILF focuses on the composition[138] of systems but not compositionality[139] to enable novel capabilities.

---

[133] Including safety and security, but also applies to all the other attributes of trustworthiness.
[134] Duan et al. *Reasoning about confidence and uncertainty in assurance cases: A survey*. Software Engineering in Health Care, 17, 2014.
[135] Bloomfield, Littlewood, Wright. *Confidence: Its role in dependability cases for risk assessment*. Annual IEEE/IFIP International Conference on Dependable Systems and Networks, 2007.
[136] NATO Science and Technology Organization, Neuilly-sur-Seine. *Framework for Semantic Interoperability*. TR-IST-094.5, 2014.
[137] Ford et al. *Purpose-aware interoperability: The ONISTT ontologies and analyzer*. Simulation Interoperability Workshop, 2007.
[138] Requiring the preservation of local properties.
[139] Compositionality requires the analysis of emergent properties of compositions, some of which are vital, as in safety and security.

More recently, self-integration based on contract theory and negotiation has been used to purposefully self-integrate, for example, drones and wearable (IoT) devices. A trust negotiation protocol for IoT devices has been developed to create an assume–guarantee contract that also includes a set of assessment procedures.[140] The contract yields additional assurance for dynamic integration from a shared historical record of adaptation assessment. This additional assurance might also be managed using the concept of assurance cases.

Other examples for self-integrating systems include mobility scenarios in which cars and, say, traffic lights are purposefully interacting and adjusting their behavior to improve the flow of traffic, and intensive care unit scenarios in which, say, heart-lung machines and X-ray cameras recognize each other and negotiate their safe interaction.[141] Some of these integrated systems could, of course, be readily constructed as bespoke *one-trick-pony* systems by suitably skilled teams. Automated self-integration, however, promises to be more flexible, more efficient, and less error prone. The above scenarios also demonstrate that, beyond automation of the integration, the challenge is to provide assurance for the safety of the integrated systems. Forming intent-driven federations of increasingly autonomous embodied systems is a challenging endeavor. Indeed, the composition of more traditional systems can often introduce new vulnerabilities,[142] as in, say, exposed crypto keys and privacy violations. We, therefore, need to come up with suitable architectural principles and composition operators for constructing resilient and safe embodied systems from a (possibly dynamically changing) set of heterogeneous and even untrusted constituent systems. In this way, embodied systems may tolerate certain failures, unexpected events, and even malicious attacks. Modeling attacks and other hardware and software defects is an issue because, almost by definition, cyberattacks are very hard to predict.[143] Yet, providing some degree of and continuously improving resilience is a must for societal acceptance of embodied systems. For the most advanced kinds of systems, what may be needed is an agreement on a shared system of ethics.

Embodied systems are acting in the real world with their wickerwork of societal norms, rules, and laws. *Smart contracts* are a central concept toward intent-driven dynamic federations of embodied systems. In this way, self-integration and self-orchestration might be approached as follows.

---

[140] Riley et al. *Toward a negotiation framework for self-integration*. IEEE International Conference on Autonomic Computing & Self-Organizing Systems Companion, 2020.
[141] Rushby. *Trustworthy self-integrating systems*. International Conference on Distributed Computing and Internet Technology, 2016.
[142] Neumann. *How might we increase trustworthiness?* Communications of the ACM, 2019.
[143] Dutertre et al. *Intrusion-tolerant enclaves*. IEEE Symposium on Security and Privacy, 2002.

- Software-based ("smart") contracts define the service interfaces and service-level agreements for embodied actors.
- Federations of embodied actors are formed through the conclusion of contracts, for instance, through bidding in auctions and/or using a mediator.
- Smart contracts are executed until the purpose of the contract has been met.

For example, a "ship" embodied actor offers smart contracts for "shipping A from B to C in exchange for D." If there is a "customer" who needs to ship "a in A" from "b in B" to "c in C" and is willing to provide "d in D" in exchange, then the "ship" and the "customer" might want to conclude a corresponding contract. If "c noting C," for example, that is the "ship" did not intend to call at harbor "c," then the "ship" might be willing to change their route for a small extra fee. Alternatively, mitigations of common mishaps might already be defined in the initial set of contracts. This is, of course, just how the current contract-based economy is designed to work. The offering and conclusion of contracts may be realized, for instance, by means of distributed execution of logic programs. Global invariants need to be maintained on a set of contracts. For example, the federation in Figure 3 needs to ensure that their mutual service-level agreements enable timely and orderly delivery of the customer's order. Other invariants need to be ensured, such as regulatory rules, desiderata, such as climate neutrality, or resiliency to some breach of contract.

Not all contracts are being served to completion. What happens, for example, if the "ship" is not able to fulfill the contract because it got stuck in some shipping channel? The federation of actors in Figure 3 may need to reorganize to still be able to satisfy the customer order on time. Such a reorganization of the federation is based on successful renegotiation and cancellation of contracts. This process of resilient execution of smart contracts generalizes the fault detection, isolation, and recovery (FDIR) cycle of fault-tolerant systems. Again, if contracts are negotiated by means of distributed logic programs, then resilient execution and renegotiation of contracts might, for example, be realized through backtracking and mechanisms for distributed incremental maintenance.

A contract-based reconfiguration of a federation might also involve a change of the *embodiment* of certain services. For example, if the ship refuses or is unable to call port "c," then the delivery federation might decide to replace this ship with other means of transportation. Using the slang of model-driven design, such a replacement involves changing the *deployment* of a *platform-independent model* (federation of virtualized services) to a *platform-dependent model* (embodiment of virtualized services in the physical world).

Most importantly, actors and/or federations of actors need to be incentivized to honor contracts. They need to be held *responsible*[144] for breach of contract possibly by some empowered higher instance, who identifies, collects evidence, and penalizes breach of contract.

Clearly, the suggested contract-based composition and execution operator for embodied systems mimics a contract-based societal organization. The obvious question is, if and how these "smart contracts" may be integrated into existing judicial systems or variants thereof. These considerations point to a multitude of serious systems programming challenges. For instance, how do we specify smart contracts? What is the right framework for negotiating contracts? How can we verify smart contracts? How can we provide evidence of the conclusion or breach of contract? How can we incentivize/penalize embodied actors to ensure beneficial behavior? Whether such federations should be deployed in social contexts without an orchestrating higher instance is also open to discussion.

## Analysis

Analysis is the process of assuring that a system meets a set of given requirements. The verification challenge involves identifying what kind of properties are expected of embodied systems and how to establish them. Analysis of embodied systems is challenging for their openness, adaptivity, situatedness, and for their largely unpredictable behavior in uncertain operating contexts. In addition to functional correctness, performance, dependability, and safety requirements as in classical embedded systems, the analysis of embodied systems focuses on establishing properties on their lawful, ethical, and robust, that is, trustworthy behavior. The verification of learning-enabled components of embodied systems poses yet another challenge.

Consider, for example, the embodied retail system in Section 1, and assume that, in reaction to the customer's request, a dynamic federation of production and logistics services have been set up to cooperatively serve this request. Analysis might establish that this federation actually is able to fulfill the request within the agreed time frame, that the delivered product is consistent with functional, quality, and safety agreements. The federation might also be shown to be robust to common faults, such as breakdowns of logistics chains and even malicious attacks. Moreover, the analysis may be used to demonstrate that the federation complies with applicable laws (for example, tax laws) and that certain social values, such as climate neutrality, are adhered to.

---

[144] Vazdanan et al. *Responsibility research for trustworthy autonomous systems*. Autonomous Agents and Multiagent Systems, 2021.

State-of-the-practice techniques for safety analysis require deterministic behavior in well-defined operating contexts and usually rely on fallback mechanisms to a human operator. Clearly, these prerequisites are not fulfilled for embodied systems and consequently, current safety analysis methodology, as encoded in industrial standards, such as DO 178C in aerospace or ISO 26262 in the automotive domain, are not applicable—at least not directly so.

Three main analysis techniques exist, namely, *testing*, *symbolic verification*, and *runtime verification*. We briefly describe some associated challenges when applied to analyzing embodied systems with learning-enabled components.

Testing. This is the most widely used and arguably also the most successful technique for analyzing software-intensive systems. Nondeterministic systems, however, are usually considered untestable because of the overwhelming number of cases to be considered. System tests are also performed with the assumption of fixed and well-described operating contexts. Embodied systems, however, need to be analyzed with respect to uncertain operating contexts, which may not even be known at design time. Finally, the analysis for learning-enabled components requires establishing properties for all possible evolutions of such a component. For all these reasons, testing methodologies as developed for embedded systems are not directly applicable to embodied systems. Novel approaches to testing embodied systems are urgently needed.

For example, scenario-based testing dynamically classifies relevant scenarios by means of automated clustering and generates a sufficient set of test cases from the classes thus obtained.[145] More generally, probabilistic programs might be synthesized for capturing relevant scenarios,[146] because probabilistic programs assign distributions to features of scenarios, and they impose hard and soft constraints over scenarios.

Testing is usually decomposed into testing individual components of a system followed by testing the integrated system. But then, how can we test systems with learning-enabled components? For ANNs, traditional structural coverage criteria from software testing can usually not be applied directly to ANN. For example, neuron coverage is trivially fulfilled for an ANN using a single test case. Moreover, branch coverage, when applied to ANNs, may lead to an exponential (in the number of neurons) number of branches to be investigated and is, therefore, not practical, as typical ANNs comprise millions of neurons. As usual, in testing, the balance between the ability to find bugs and the computational cost of test case generation is essential for the effectiveness of a test method.[147]

---

[145] Hauer. *On Scenario-Based Testing of Automated and Autonomous Driving Systems*. Doctoral dissertation, Technische Universität München, 2021.
[146] Fremont et al. *Scenic: Language-Based Scene Generation*. Technical Report No. UCB/EECS-2018-8, 2018.

Therefore, ANN-specific nonstructural test coverage criteria for the robustness, interpretability, completeness, and correctness of an ANN have been developed.[148] A scenario coverage metric, for example, partitions the possible input space according to N attributes (for example, snow and rainy), and proposes, on the basis of existing work on combinatorial testing, efficient k-projection (for k = 0, …, N–1) coverage metrics as approximations of the exponential number of input partitions.

The generation of falsifying/adversarial test cases generally uses search heuristics based on gradient descent or evolutionary algorithms.[149] [150] [151] [152] These approaches may be able to find falsifying examples efficiently, but they usually do not provide an explicit level of confidence about the nonexistence of adversarial examples in case the algorithm fails to find one.

Various traditional techniques for test case generation, such as fuzzing, symbolic execution, concolic testing, mutation testing, and metamorphic testing, have been extended to ANNs. Despite their effectiveness in discovering various defects of ANNs together with their data-centered requirement specifications, however, how testing-based approaches can be efficiently integrated into the construction of convincing safety argumentations for learning-enabled components, let alone embodied systems, is not exactly clear.

Altogether, testing methods seem to be effective at discovering defects of learning-enabled components such as ANNs. However, how to measure the effectiveness of test coverage metrics in building up sufficient confidence or, dually, raising doubts is unclear. Moreover, most testing-based approaches assume a fixed ANN. Yet, ANNs are learning-enabled and trained continuously on new data/scenarios. The challenge is to come up with methodologies for efficiently—depending on the application context also in real time—retesting safety requirements for continuously evolving ANNs.

Symbolic Analysis. These analysis techniques generalize testing, in that, sets of test cases are evaluated on a system at once. These test sets are usually encoded as logical constraints for describing possibly infinite test sets. Symbolic analysis requires neither a complete system implementation nor a fully specified operational context, because unknown behavior may be represented logically by means of uninterpreted functions. Logical constraints on these uninterpreted functions are used to express known (or learned) facts about these behaviors. In contrast to testing, a symbolic analysis may be applied to demonstrate that certain requirements hold for embodied systems that operate in uncertain and only partially known operating contexts. Another use of symbolic analysis is to support the generation of safe trajectories during runtime.[153]

---

[147] Sun et al. *Testing deep neural networks*. arXiv:1803.04792v4, 2019.
[148] Cheng, Nührenberg, Rueß, Yasuaoka. *Towards dependability metrics for neural networks*. 16th ACM/IEEE International Conference on Formal Methods and Models for System Design, 2018.
[149] Goodfellow, Shlens, Szegedy. *Explaining and harnessing adversarial examples*. arXiv:1412.6572, 2014.
[150] Papernot et al. *The limitations of deep learning in adversarial settings*. IEEE European Symposium on Security & Privacy, 2016.
[151] Carlini, Wagner. *Towards evaluating the robustness of neural networks*. IEEE European Symposium on Security & Privacy, 2017.

Recently, many different symbolic analysis techniques have been adapted to learning-enabled components such as ANNs.[154] Verification problems for ANNs have been reduced to constraint solving problems such as satisfiability in propositional logic,[155] [156] satisfiability modulo theories,[157] [158] [159] [160][161] and mixed-integer linear programming.[162] These approaches, however, typically do not scale up to the size of real-world ANNs with millions of neurons. Approximation techniques are applied to improve efficiency but usually at the expense of precision. Recent approaches based on global optimization have the potential to deal with larger networks.[163] Compositional verification techniques for scaling up symbolic analysis are largely missing. Symbolic analysis technologies work on abstract models, which is why they might miss certain defects due to implementation issues (for example, rational numbers vs. IEEE floating points). Moreover, how to efficiently apply these techniques to continuously evolving learning-enabled components is also unclear.

Runtime Verification and Recovery. In runtime verification, a monitor observes the concrete execution of the system in question and checks for violations of stipulated properties. When the monitor detects a violation of a property, it notifies a command module, which then isolates the cause of the violation, followed by an attempt to recover from the violation. In this way, runtime verification is a central element of FDIR-based[164] fault-tolerant systems. For the multitude of sources for uncertainty in AI systems, stringent real-time requirements, and ever-changing learning-enabled components, runtime verification is an essential element for analyzing embodied systems.

---

[153] Althoff, Dolan. *Online verification of automated road vehicles using reachability analysis*. IEEE Transactions on Robotics, 2014.
[154] Huang et al. *A survey of safety and trustworthiness of deep neural networks: Verification, testing, adversarial attack and defence, and interpretability*. Computer Science Review, 37, 2020.
[155] Cheng, Nührenberg, Rueß. *Verification of binarized neural networks*. Working Conference on Verified Software: Theories, Tools, and Experiments, 2018.
[156] Narodytska et al. *Verifying properties of binarized deep neural networks*. Proceedings of the AAAI Conference on Artificial Intelligence, 2018.
[157] Huang, Kwiatkowska. Safety verification of deep neural networks. International Conference on Computer Aided Verification, 2017
[158] Pulina. *An abstraction-refinement approach to verification of artificial neural networks*. International Conference on Computer Aided Verification, 2010.
[159] Katz et al. *Reluplex: An efficient SMT solver for verifying deep neural networks*. International Conference on Computer Aided Verification, 2017.
[160] Tuncali, Ito, Kapinski, Deshmukh. *Reasoning about safety of learning-enabled components in autonomous cyber-physical systems*. Proceedings of the 55th Annual Design Automation Conference, 2018.
[162] Cheng, Nührenberg, Rueß. *Maximal resilience of artificial neural networks*. International Symposium on Automated Technology for Verification and Analysis, 2017.
[163] Ruan, Wu, Sun, Huang. *Reachability analysis of deep neural networks with provable guarantees*. IJCAI, 2018.
[164] Fault Detection, Isolation, and Recovery

Architectural design principles for monitoring distributed systems are needed to ensure that monitoring does not perturb the system (at least, not too much).[165] In particular, a tutorial discusses challenges involved in instrumenting real-time systems so that the timing constraints of the system are respected.[166] A recent tutorial describes state-of-the-practice technology for generating runtime monitors that capture the safe operational environment of systems with AI/ML components.[167]

Altogether, runtime verification is an essential and attractive technique of any verification strategy for embodied systems. Unlike static verification techniques, such as testing or symbolic analysis, adaptation to learning-based components is not needed. In this way, runtime monitoring is an enabling verification technology for continuous assurance based on the MAPE-K[168] loop from autonomic computing. The main challenge in deploying runtime monitoring, as is the case for any other CPS, is to embed monitors in an efficient (for example, energy-efficient) way without perturbing the behavior of the embodied system too much.

Runtime monitoring may also be used to measure uncertainties in the input–output behavior of learning-enabled components. For example, if the input is out of the distribution of the training set, then one may conclude that the output may not actually be a "correct" one. Such information about the uncertainty of a perception result is useful input for deliberatively planning meaningful and safe actions. Uncertainty information about the perception unit is also used in Simplex architectures for switching to a safe(r) perception channel whenever the ANN output is doubtful. Clearly, the distance (in some given metric) of the input to the set of training input may serve as a measure of uncertainty of the input–output behavior of the learning-enabled component. However, such a measure returns zero uncertainty even for "incorrect" behavior on training inputs. Alternatively, a proposed approach is to monitor the neuronal activation pattern in ANN-based components and to compare it with neuronal activation patterns as learned during the ANN training phase.[169] Applicable background knowledge and physical laws may also be used in monitoring the plausibility of the input-output behavior of an ANN.

---

[165] Goodloe, Pike. *Monitoring distributed real-time systems: A survey and future directions*. NASA, 2010.
[166] Bonakdarpour. *Runtime-monitoring of time-sensitive systems*. International Conference on Runtime Verification, 2011.
[167] https://uva-mcps-lab.github.io/RV21/paper10.1.html
[168] Measure, Analyze, Plan, Execute; the K stands for Knowledge
[169] Cheng, Nührenberg, Yasuoka. *Runtime monitoring neuron activation patterns*. IEEE Design, Automation & Test in Europe Conference & Exhibition, 2019.

In summary, due to the multitude of sources of uncertainty of embodied systems with learning-enabled components and the partially unknown environments in which they operate, even if all the challenges for specification and verification are solved, it is likely that one will not be able to prove unconditional safe and correct operation. Situations will always exist in which we do not have a provable guarantee of correctness. Therefore, techniques for achieving fault tolerance and error resilience at runtime must play a crucial role.

However, a systematic understanding does not yet exist as to what kind of analysis can be achieved at design time, how the design process can contribute to the safe and correct operation of the embodied system at runtime, and how the design time and runtime analysis techniques can interoperate effectively.

The distributed and dynamic nature of federations of embodied actors and their goals is particularly challenging for runtime verification. A runtime monitoring framework for embedded systems must support reasoning under uncertainty[170] [171] and also partially observable systems with nondeterministic and probabilistic dynamics.[172]

VerifAI is a runtime monitoring framework for autonomous systems with learning-enabled components.[173] It includes formal modeling of the autonomous system and its environment (in terms of probabilistic programs), automatic falsification of system-level specifications and other simulation-based verification and *testing* methods, automated diagnosis of errors, and automatic specification-driven parameter and component synthesis.

The safety of systems with learning-enabled components in simplex architectures[174] often relies on a runtime monitor-based switch between a performant and a safe channel. Runtime monitoring of typical security hyperproperties,[175] [176] privacy policies,[177] and contextual integrity[178] have also been considered. Given a set of properties and an embodied system, the challenge is to generate and maintain sound and possibly complete runtime monitors, which are woven into the embodied system, to keep interferences with the core behavior at a minimum.

---

[170] Zheng, Julien. Verification and validation in cyber physical systems: Research challenges and a way forward. IEEE Software Engineering for Smart Cyber-Physical Systems, 2015.
[171] Ma et al. *Predictive monitoring with logic-calibrated uncertainty for cyber-physical systems*. ACM Transactions on Embedded Computing Systems, 2021.
[172] Viswanadha et al. Parallel and multi-objective falsification with scenic and VerifAI. International Conference on Runtime Verification, 2021.
[173] Torfah, Junges, Fremont, Seshia. *Formal analysis of AI-based autonomy: From modeling to runtime assurance*. International Conference on Runtime Verification, 2021.
[174] Desai, Ghosh, Seshia, Shankar, Tiwari. *SOTER: A runtime assurance framework for programming safe robotics systems*. 49th Annual IEEE/IFIP International Conference on Dependable Systems and Networks, 2019.
[175] Finkbeiner, Hahn, Stenger, Tentrup. *Monitoring hyperproperties*. Formal Methods in System Design, 54(3), 2019.
[176] Bonakdarpour, Sanchez, Schneider. *Monitoring hyperproperties by combining static analysis and runtime verification*. International Symposium on Leveraging Applications of Formal Methods, 2018.
[177] Chowdhury, Jia, Garg, Datta. *Temporal mode-checking for runtime monitoring of privacy policies.* International Conference on Computer Aided Verification, 2014.
[178] Barth, Datta, Mitchell, Nissenbaum. *Privacy and contextual integrity: Framework and applications*. IEEE Symposium on Security & Privacy, 2006.

## Assurance

A trustworthy embodied system operating in our very economic and social fabric is required to obey applicable regulations and laws, to act according to human-centered social values, and to be robust and safe (see Section 3). But how can we be assured that an embodied system is indeed worthy of the trust we may put in it? Clearly, an analysis (see Subsection Analysis) of embodied systems is a central component of any mechanism for inspiring confidence in the lawful, ethical, and robust behavior of any technical system, and for building up trust. Additionally, as an actor in a real-life context, an embodied system is required to provide explicit evidence that it is acting as required.

Requirements. Embodied systems have the possibility of autonomously acting in regulated sectors such as healthcare, finance, insurance, accounting, or retail. As such, they need to comply with applicable regulations and national laws, for example, the General Data Protection Regulation, Health Insurance Portability and Accountability Act, the Gramm–Leach–Bliley Act, the Sarbanes–Oxley Act, or the Children's Online Privacy Protection Act.

These kinds of regulations and laws are indeed not lacking. For example, German law now obliges organizations to demonstrably ensure that basic labor and human rights standards are respected throughout global supply chains.[179] The length of these regulations and laws, the opacity of the legal language, the complexity of these acts, and their contradictions,[180] however, make determining and demonstrating compliance difficult.[181]

Regulations have been formalized using a variety of policy languages.[182] [183] [184] Formalisms and systems for expressing and checking privacy policy include P3P,[185] EPAL,[186] XACML,[187] DKAL,[188] and the logic of privacy and utility, which formalizes the concept of *contextual integrity*.[189] In addition to compliance with applicable laws and norms, embodied systems are required to be demonstrably *robust*, *safe,* and *secure*.

---

[179] https://www.bmz.de/de/entwicklungspolitik/lieferkettengesetz
[180] Lam, Mitchell, Sundaram. *A formalization of HIPAA for a medical messaging system*. International Conference on Trust, Privacy and Security in Digital Business, 2009.
[181] Ness. *A year is a terrible thing to waste: early experience with HIPAA*. Annals of Epidemiology, 2(15), 2005.
[182] Anton, Earp, Reese. *Analyzing website privacy requirements using a privacy goal taxonomy*. Proceedings IEEE Joint International Conference on Requirements Engineering, 2002.
[183] Anton, Eart, Vail, Jain, Gheen, Frink. *HIPAA's effect on web site privacy policies*. IEEE Symposium on Security & Privacy, 2007.
[184] Anton, He, Baumer. *Inside JetBlue's privacy policy violations*. IEEE Symposium on Security & Privacy, 2004.

More precisely, embodied systems are expected to be

- resilient to common and possibly also new kinds of breakdowns and malicious attacks,
- such that the risk of unintended harm to humans, machinery, and the environment is demonstrably below acceptable levels, and
- such that identified confidentiality, integrity, and availability requirements are satisfied.

Other requirements include *transparency* (for example, traceable use of data and information sources, published decision policies), demonstrable *fairness* of decisions (for example, all applicants are demonstrably being served equally), *inverse privacy*,[190] or *contextual integrity*.[191] [192] The latter requirement is based on the hypothesis that privacy is a right to appropriate flow of information. In case of noncompliance with any of these requirements, culprits need to be identified and called to account.

The challenge is to come up with satisfactory formalizations of adequate subsets of applicable laws and norms as a prerequisite for demonstrable compliance with these requirements. Previous attempts in this direction were often based on some variant of temporal logic and/or logic programming.

We now describe possible approaches and corresponding challenges for assuring compliance with these kinds of requirements. The challenge for a federation of embodied actors is to assure, and possibly demonstrably so, that it is indeed worthy of our trust (Section: Trustworthiness). For example, the largely autonomously acting federation of embodied retail actors (Section: Service Federations) is required to answer questions such as the following:
Is it able to deliver the requested product in time and according to order specs? Is it resilient to common planning breakdowns, failures, and malicious attacks? Are contingency plans and *never-give-up* strategies put in place in case of, say, operational, judicial, and financial quagmires? Are the actors of the federation acting safely? Does the federation obey applicable tax laws? Can the federation demonstrate that the delivered product is not based on child labor? Does the federation demonstrably act in a climate-neutral or even climate-positive manner? This list of questions is necessarily incomplete as there is indeed no lack of requirements when acting in real-world social contexts.

---

[185] Cranor. *P3P: Making privacy policies more useful*. IEEE Symposium on Security & Privacy, 2003.
[186] Ashley, Hada, Karjoth, Powers, Schunter. *Enterprise privacy authorization language (EPAL)*. IBM Research, 30, 2003.
[187] Ramli, Nielson, Nielson. *The logic of XACML*. Science of Computer Programming, 83, 2014.
[188] Gurevich, Neeman. *DKAL 2—A simplified and improved authorization language*. Technical Report MSR-TR-2009–11, 2009.
[189] Barth, Datta, Mitchell, Nissenbaum. *Privacy and contextual integrity: Framework and applications*. IEEE Symposium on Security & privacy, 2006.
[190] Gurevich, Hudis, Wing. *Viewpoint: Inverse privacy*. Communications of the ACM, 59(7), 2016.
[191] Nissenbaum. *Privacy in Context: Technology, Policy, and the Integrity of Social Life*. Stanford University Press, 2010.
[192] Datta, Blocki, Christin, DeYoung, Garg, Jia, Sinha. *Understanding and protecting privacy: Formal semantics and principled audit mechanisms*. International Conference on Information Systems Security, 2011.

Analysis-based Assurance. Static analysis techniques, such as testing, ensure the system's compliance with requirements, but only at selected points of time in their evolution. Even if one succeeds in testing the compliance of an embodied system with its requirements, these results are usually outdated in further evolutions of the self-learning embodied system operating in a dynamic context, thereby requiring retesting. In a similar vein, traditional formal verification techniques may be applied to demonstrate, *a priori* during design time, conformance requirements of an embodied system operating in largely unknown environments and all its possible evolutions. Hereby, proof objects serve as evidence,[193] and reflection principles are used to generate specific evidence for each possible evolution.[194] But it seems unrealistic that still largely manual formal verification is applied to the analysis of nontrivial evolving embodied systems.

By and large, we, therefore, need to go beyond current static analysis techniques, such as testing and formal verification, to realistically assure compliance of ever-evolving embodied systems with their complex real-world requirements.

Process-based assurance. Assurance has traditionally been approached in *software and systems engineering* through design processes that follow rigorous development standards and by demonstrating compliance through analysis. Well-proven and successful assurance standards, such as *DO 178C* in aerospace or *ISO 26262* in the automotive domain, are predicated on the assumptions that system behavior is deterministic, a clearly defined operating environment exists, a fallback to a human-in-the-loop operator is always available, and the system does not learn and evolve once it is deployed.

These basic assumptions of traditional safety engineering do not hold anymore for embodied systems. Safety engineering, therefore, comes to a grand climacteric, moving from deterministic, nonevolving embedded systems and CPSs operating in well-defined contexts to embodied systems. Process-based design for many embodied systems seems to be difficult, if not impossible; as these systems continuously integrate into dynamic coalitions, they are increasingly autonomous, and they are evolving based on previous experience. But changes in process-based assurance are notoriously slow and expensive.[195] Therefore, assurance for embodied systems increasingly needs to rely on assurance artifacts collected at runtime instead of process-based design-time assurance.

---

[193] Necula. *Proof-carrying code*. Proceedings of the 24th ACM SIGPLAN-SIGACT Symposium on Principles of Programming Languages, 1997.
[194] Rueß. *Meta-Programming in the Calculus of Construction*. Doctoral dissertation, Ph.D. thesis, Universit at Ulm, 1995.

Continuous Assurance. The general recommendation is to audit the trustworthiness of an embodied system continuously throughout its life cycle (Table 1) to assure compliance with applicable laws, social norms, and technical robustness and safety requirements. This is easier said than done, as embodied systems

- are dynamic federations based on changing priorities and intent,
- are composed of ever-evolving embodied systems with self-learning capabilities, and
- operate in largely unknown contexts based on experience and targeted exploration.

Moreover, applicable laws and regulations may change over time and are also location-dependent.

The challenge, therefore, is *continuous assurance*, an assurance system that is provided provisionally at design time and continually monitored, updated, and evaluated at runtime as the system and its environment evolves. But continuous assurance for regulations, such as GDPR, has become a costly burden.[196] [197] The culprit is the heavy reliance on manual checklist-based auditing in today's compliance process, which is expensive, slow, and error prone. It is also restricted to the compliance of static snapshots in the evolution of a system. Therefore, runtime analysis and assurance are important for continuously assuring compliance of ever-evolving embodied systems.

*Runtime assurance* is a dynamic analysis technique based on a log for recording system events, which are relevant for establishing conformance with given requirements. These requirements

---

[195] Changes of only one line of code in safety-critical code in the aerospace industry have reportedly cost seven-figure sums (in US dollars) in some cases.
[196] According to Forbes, GDPR cost US Fortune 500 companies $7.8 billion as of 2018. (Forbes. *The GDPR racket: Who's making money from this $9bn business shakedown*. 2020. https://www.forbes.com/sites/oliversmith/2018/05/02/the-gdpr-racket-whos-making-money-from-this-9bn-business-shakedown/#54c0702034a2)
[197] A recent report shows that 74% of small or midsized organizations spent more than $100,000 to prepare for continuous compliance with GDPR and CCPA. (DataGrail. *The age of privacy: The cost of continuous compliance*. 2020. https://datagrail.io/downloads/GDPR-CCPA-cost-report.pdf)

Systems Challenges for Trustworthy Embodied Systems                                                    55

are usually expressed in a variant of temporal logic, which is expressive enough for encoding regulation rules.[198] Runtime verification techniques are used to establish that the logged event trace indeed conforms to the given requirement. For example, an embodied actor logs events of its data handling operation, and runtime verification is used to establish compliance with a given privacy policy or detect violations thereof. The root cause of such a violation may be computed from a log as the basis for calling the responsible embodied actors to account for the conformance violations they caused.

A subject that is open to discussion is if the identified embodied culprits can and should be treated as legal entities that can be held responsible for their actions in a traditional brick-and-mortar judicial system. Workable judicial governance of one sort or the other is needed for promoting, by and large, compliant behavior as the basic mechanism for self-improving system behavior.

Depending on the regulatory framework, compliance with these requirements by every actor might lead to functionally inferior behavior or may even be a barrier to innovation. For example, regulatory frameworks may well be contradictory; considerations of the relative importance of compliance with a wide range of different requirements, possibly from different sources, are ongoing; and, finally, without minor transgressions of, say, traffic rules (such as crossing solid lines), traffic flow would often be less fluid.

At a high-level, runtime assurance can be approached by compliance checks on recorded logs when demanded, say, by an audit authority, [199] by online checking of relevant events against the prevailing policy, [200] [201] [202] [203] [204] [205] [206] [207] or by combinations thereof.[208] In federations of

---

[198] DeYoung, Garg, Dia, Kaynar, Datta. *Experiences in the logical specification of the HIPAA and GLBA privacy laws*. Proceedings of the 9th Annual ACM Workshop on Privacy in the Electronic Society, 2010.
[199] Garg, Jia, Datta. *Policy auditing over incomplete logs: Theory, implementation and applications*. Proceedings of the 18th ACM Conference on Computer and Communications Security, 2011.
[200] Basin, Klaedtke, Müller. *Monitoring security policies with metric first-order temporal logic*. Proceedings of the 15th ACM Symposium on Access Control Models and Technologies, 2010.
[201] Basin, Klaedtke, Marinovic, Zalinescu. *Monitoring compliance policies over incomplete and disagreeing logs*. International Conference on Runtime Verification, 2012.
[202] Basin, Klaedtke, Marinovic, Zalinescu. *Monitoring of temporal first-order properties with aggregations*. International Conference on Runtime Verification, 2013.
[203] Chomicki. *Efficient checking of temporal integrity constraints using bounded history encoding*. ACM Transactions on Database Systems, 20(2), 1995.
[204] Chomicki, Niwinski. *On the feasibility of checking temporal integrity constraints*. Journal of Computer and System Sciences, 51(3), 1993.
[205] Krukow, Nielsen, Sassone. *A logical framework for history-based access control and reputation systems*, Journal of Computer Security, 16(1), 2008.
[206] Bauer, Gore, Tiu. *A first-order policy language for history-based transaction monitoring*. International Colloquium on Theoretical Aspects of Computing, 2009.
[207] Ozeer. φ comp: An architecture for monitoring and enforcing security compliance in sensitive health data environment. 18th International Conference on Software Architecture Companion, 2021.
[208] Chowdhury, Jia, Garg, Datta. *Temporal mode-checking for runtime monitoring of privacy policies*. International Conference on Computer Aided Verification, 2014.

embodied actors, system logs not only need to contain all relevant events in the right order but also need to be tamper-proof and to comply with applicable privacy policies. Distributed ledger technology, such as various incarnations of blockchain, has lately been tried, for instance, for tamper-proof logging. Monitoring information flow policies such as noninterference or observational determinism, which relate multiple computation traces with each other, is also challenging.[209] [210]

Assurance Cases. A fundamentally different approach to assurance is based on constructing and maintaining safety and assurance cases,[211] which are compelling, comprehensive, defensible, and valid justifications of the compliance of a system. It is based on a structured argument of assurance considerations across the system life cycle that can assist in convincing various stakeholders that the system is acceptably safe. The purpose is, broadly, to demonstrate that the safety-related risks associated with specific system concerns[212] have been identified, are well understood, and have been appropriately mitigated, and that mechanisms are in place to monitor the effectiveness of safety-related mitigations. In this sense, an assurance case is a structured argument for linking safety-related claims through a chain of arguments to a body of the appropriate evidence.

One of the main benefits of structured arguments in assurance cases is that the causal dependencies between assurance claims and the substantiating evidence as obtained, for instance, by analysis are captured explicitly. Assurance cases also determine the level of scrutiny needed for developing and operating systems that are acceptably safe. This kind of information from safety cases might also be used by an embodied actor in the future to safely explore and navigate in largely unknown operating contexts.
However, assurance cases currently are at most semiformal, and their construction and maintenance require significant manual efforts. Adequate notions of assurance case patterns and modular composition operators for assurance cases are opening up new possibilities for aligning compliance arguments with system evolutions during runtime (such as the dynamic formation of federations).

---

[209] Finkbeiner, Hahn, Stenger, Tentrup. *Monitoring hyperproperties*. International Conference on Runtime Verification, 2017.
[210] Bonakdarpour, Finkbeiner. *The complexity of monitoring hyperproperties*. IEEE 31st Computer Security Foundations Symposium, 2018.
[211] UK Ministry of Defence, 2007.
[212] Including safety and security, but also applies to all other attributes of trustworthiness.

A major challenge is to measure configdence in assurance cases.[213] [214] [215] [216] [217] Eliminative induction increases confidence in assurance cases by removing sources of doubt and using Baconian[218] probability to represent confidence.[219] One systematic approach is the construction and dialectical consideration of counterclaims and countercases, where counterclaims are natural in *confirmation measure*s as studied in Bayesian confirmation theory, and countercases are assurance cases for negated claims.

Assurance cases are successfully applied to traditional safety-critical systems with clearly defined operating contexts, safety requirements, and fallback strategies to human operators. Given the increasing complexities and sources of uncertainty, however, the current assurance approach with prescribed and fixed verification and validation process activities, criteria, and metrics does not work well for assuring AI-based or even embodied systems. With the multitude of sources of uncertainty, assurance arguments for increasingly autonomous embodied systems need to (1) stress rigor in the assessment of the evidence and reasoning employed and (2) automate the search for defeaters, the construction of cases and countercases, and the management and representation of dialectical examination.[220] A topic of particular interest is to capture how the influence of a learning-enabled component is depicted and reasoned within the control structure of an embodied actor. Rigorous assurance cases can be adapted faster and more efficiently to the ever-evolving assurance needs of embodied systems.

Rigorous assurance cases open new possibilities for dependable and safe exploration in largely unknown operating contexts, which have been obtained from relevant information of a safety case and its certainty assessment. For example, if only weak evidence is available for the fact that the traffic light in front of the *ego* vehicle is green, then the ego cart might want to increase its assurance by strengthening this case through additional sensor activity, for example. Therefore, rigorous assurance cases can be instrumental in online behavioral self-adaptation and in determining safe behavior when operating in uncertain contexts.

---

[213] Grigorova, Maibaum. *Taking a page from the law books: Considering evidence weight in evaluating assurance case confidence*. IEEE International Symposium on Software Reliability Engineering Workshops. 2013.
[214] Duan, Rayadurgam, Heimdahl, Sokolsky, Lee. *Representing confidence in assurance case evidence*. International Conference on Computer Safety, Reliability, and Security, 2014.
[215] Bloomfield, Littlewood, Wright. *Confidence: Its role in dependability cases for risk assessment*. IEEE 37th Annual IEEE/IFIP International Conference on Dependable Systems and Networks, 2007.
[216] Rushby. *Formalism in safety cases*. Making Systems Safer, Springer, 2010.
[217] Rushby. *Logic and epistemology in safety cases*, International Conference on Computer Safety, Reliability, and Security, 2013.
[218] https://ntrs.nasa.gov/api/citations/20160013333/downloads/20160013333.pdf
[219] Goodenough, Weinstock. *Toward a theory of assurance case confidence*. CMU Technical Report, 2004.
[220] Bloomfield, Rushby. *Assurance 2.0: A manifesto*. arXiv:2004.10474, 2020.

*Evidence-based assurance* is based on explicitly generating verifiable evidence for any transaction of and between embodied actors. The underlying mindset is "verify, then trust." Consider a responsible visa-granting actor that returns evidence (think of it as a number) corresponding to the permission for the requester to enter the foreign country for a specific time frame. The border control verifies the identity of the presenter upon entry and verifies that the presented visa is valid. This basic mechanism is applicable to all kinds of transactions that involve the exchange of physical evidence in the form of identity cards, driver's licenses, money, checks, visas, airline tickets, traffic tickets, birth certificates, vaccination certificates, and stock certificates, as well as electronic evidence, including PIN numbers, passwords, keys, certificates, digital coins, and nonces.

It is easy to extrapolate from the above scenario the other uses of digital evidence in electronic commerce, business and administrative processes, and digital government, where actors execute smart contracts (as outlined in the Self-Integration subsection) to perform specific tasks that require the exchange of authorization and authentication information. In this way, evidence is produced, say, to assert the conclusion of contracts. The generated evidence can then be independently and automatically checked as the prerequisite for directing corresponding payments.

Previously, we defined a framework for evidential transactions called Cyberlogic, which is based on public key infrastructure.[221] [222] The basic components are simple. First, evidence is encoded by means of numbers using digital certificates and nonces. Second, predicates are signed by private keys so that decryption of such a certificate with the corresponding public key serves as proof or evidence for the assertion contained in the certificate. Third, protocols are distributed logic programs that gather evidence using ordinary predicates, digital certificates, and proof construction based on the Curry–Howard isomorphism. These simple building blocks are sufficient to encode all-important capabilities for delegation, retraction of permission, and time-stamped evidence. For example, some vaccination certificates might be handed out at a certain point in time. Now, new medical evidence on the ephemeral efficacy of vaccination results in the need to retract corresponding capabilities.

Evidential transaction in a Cyberlogic-like setting has recently been used to demonstrate that accountability for federated ML is paramount to fully overcoming legislative and jurisdictional constraints.[223] In particular, it ensures that all entities' data are adequately included in the model and evidence on fairness and reproducibility is curated toward trustworthiness or realizing accountability for federated ML. Cyberlogic also forms the foundations for secure and trusted-by-design smart contracts.[224] Even though Cyberlogic has proven to be instrumental in many cases, whether proof-carrying capabilities can be scaled up for complex real-world systems is still unclear. Possible directions for addressing these scalability issues are interactive proof systems and the use of advanced cryptographic mechanisms.

---

[221] Rueß, Shankar. *Introducing cyberlogic*. Proceedings of the 3rd Annual High Confidence Software and Systems Conference, 2003.
[222] Nigam, Reis, Rahmouni, Rueß. *Proof search and certificates for evidential transactions*. Automated Deduction–CADE 28, 2021.
[223] Balta, Sellami, Kuhn, Schöpp, Buchinger, Baracaldo, Altakrouri. *Accountable federated machine learning in government: Engineering and management insights*. International Conference on Electronic Participation, 2021.

We have described challenges and possible approaches for the assurance of embodied systems acting in social contexts. A detail that should have become clear is that embodied systems, as increasingly autonomous actors in the real world, need to deal, like the rest of us, with the complexity and rigmaroles associated with real-world norms and laws. As technical systems, however, they rely on rigorous and faithful digital representations thereof. Moreover, auditing, as a means of assuring compliance, needs to be aligned with the evolvement of actors. Auditing also needs to be largely automated. Correspondingly, we have outlined promising base technologies based on assurance cases and evidential transactions.

We suggest that rigorous and automated assurance is key to a meaningful embodiment of ever-evolving and increasingly autonomous systems into social contexts. We also hope that these advancements may also eventually prove to be instrumental for more rational and evidence-based development and maintenance of judicial and regulatory frameworks as the basis for deforesting the convoluted jungle of applicable regulations and for creating space for more innovative behavior.

---

[224] Dargaye, Kirchner, Tucci-Piergovanni. *Towards secure and trusted-by-design smart contracts*. Francophone Days of Application Languages, 24, 2018.

# 5. Conclusions

We have argued that a new generation of increasingly autonomous and self-learning systems is about to be developed and embodied into all aspects of everyday life. The main driver for their deployment is their ubiquitous disruptive potential. Autonomy and unsupervised learning capabilities, therefore, are widely believed to be the key technological base for initiating and driving the next economic and societal phase shift.[225]

Embodied systems are not distant AI-ish science fiction, as purpose-built embodied systems might be "handcrafted" with currently available technology, but only at a very high cost and sometimes with unknown risks, because we do not yet have mature science and technology to support the engineering of embodied systems in which we may place our trust.

Becoming (too) dependent and losing meaningful control over increasingly autonomous and self-learning systems are real risks. When deploying these systems into real-life contexts, we face various engineering challenges, because coordinating the behavior of embodied actors in a beneficial manner is crucial to ensure their compatibility with our human-centered social values and to design verifiably safe and reliable human–machine interaction.[226]

Traditional systems engineering, however, is coming to a climacteric from embedded to embodied systems, and with assuring the trustworthiness of a new generation of dynamic federations of situationally aware, intent-driven, explorative, ever-evolving, largely unpredictable, and increasingly autonomous embodied systems in uncertain, complex, and unpredictable real-world contexts.

Model-driven engineering has so far been the pillar for building dependable and safe embedded control systems, such as those used for flying airplanes safely. Currently, we are working on data-driven engineering techniques for functional automated systems with learning-enabled components. Next, we need to develop engineering capabilities for trustworthy embodied systems, which are based on a suitable mixture, adaptations, and further developments of the model- and data-driven engineering methodologies.

Such an engineering framework needs to be as rigorous as possible, because embodied systems will eventually be equipped with substantial self-engineering capabilities, including data- and experience-driven functional updates, zero-touch repair and maintenance capabilities, and the possibility of by-need-augmentation of sensing, cognitive, and acting capabilities. We also envision embodied systems that perform their own risk analysis and define their own mitigation strategies based on their own understanding of socially acceptable behavior.

---

[225] Davidow, Malone. *The Autonomous Revolution: Reclaiming the Future we've Sold to Machines.* Berrett–Koehler Publishers, 2020.
[226] Vazdanan et al. *Responsibility Research for Trustworthy Autonomous Systems*, 2021.

We have described the main characteristics of embodied systems and derived from these characteristics a shortlist of central systems challenges, including robust and human-centered AI, architectures for autonomous systems, safe exploration of the unknown (unknown), trustworthy self-integration, and continual verification and assurance for federated self-learning systems. We restricted ourselves to technical systems challenges, thereby omitting important but largely unsolved economic, jurisdictional, and societal questions on embodying autonomous, technical systems into our everyday life.

We consider the identified list of technical systems challenges to comprise only the most pressing ones. We expect to identify additional ones as we gain a deeper understanding of the real consequences of embodying autonomous actors into economic and societal contexts. Moreover, the relative importance of each of the identified systems challenges is highly dependent on the envisioned application context and its associated requirements. Solving these challenges will require synergistic innovations in software systems engineering, architectures for autonomous systems, and core AI/ML algorithms.

A crucial next step is to gain more experience and increase our theoretical understanding of autonomous and self-learning embodied systems in which we can put our trust. A series of increasingly challenging embodied systems might support us in leveraging and bootstrapping engineering knowledge for embodied systems in an accelerated fashion. At the same time, sound sociopolitical and legal conditions and frameworks must be created to embody autonomously acting machines in essential real-world processes and structures, because failing to deploy embodied systems in a meaningful manner into our very economic and social fabric can all too easily and quickly become dystopian.

# Imprint

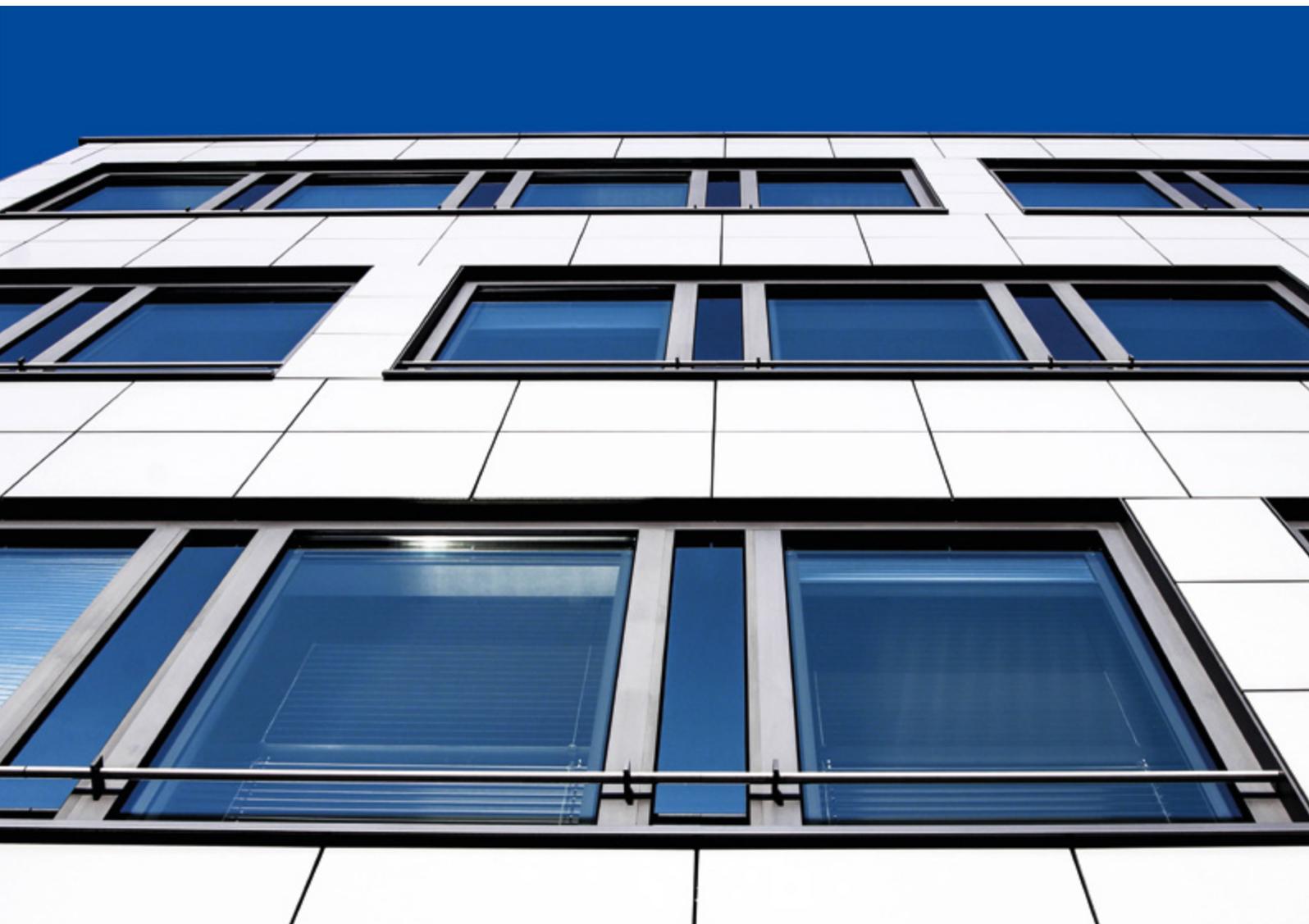

**fortiss GmbH**
Guerickestraße 25
80805 München
Deutschland
www.fortiss.org
Tel.: +49 89 3603522 0
E-Mail: info@fortiss.org

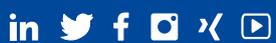

fortiss